\documentclass[journal]{IEEEtran}
\ifCLASSINFOpdf
  \usepackage[pdftex]{graphicx}
  \usepackage{subcaption}
  \usepackage{caption}
  \usepackage{subfiles} 
  \usepackage{multirow}
  \usepackage[font={footnotesize}]{caption}
  \usepackage[hyphens]{url}
  \usepackage[colorlinks=true, allcolors=black, breaklinks=true, urlcolor=blue]{hyperref}
  \usepackage{xcolor}
  \usepackage{booktabs}
  \usepackage{algorithm}
  \usepackage{algpseudocode}
  \usepackage{tcolorbox}
\else
\fi
\usepackage{amsmath}
\usepackage{amssymb}
\usepackage{pifont}

\DeclareRobustCommand{\tick}{\ding{51}}
\DeclareRobustCommand{\xmark}{\ding{55}}
\begin{document}
%
\title{Closing the Loop: Unified 3D Scene Generation and Immersive Interaction via LLM–RL Coupling}

%
%

\author{Anh H. Vo,~\IEEEmembership{}
        Sungyo Lee,~\IEEEmembership{}
        Phil-Joong Kim,~\IEEEmembership{}
        Soo-Mi Choi,~\IEEEmembership{}
        and Yong-Guk Kim*~\IEEEmembership{}
        
\thanks{A H Vo, S Lee, P-J Kim, S-M Choi, and Y-G Kim are with the Department of Computer Engineering, Sejong University, Seoul, Republic of Korea.}

\thanks{* Corresponding Author: ykim@sejong.ac.kr.}

} 

%
%

\markboth{Journal of \LaTeX\ Class }%
{Shell \MakeLowercase{\textit{et al.}}: Bare Demo of IEEEtran.cls for IEEE Journals}
%



\maketitle
Recent advances in large language models (LLMs) have significantly improved language-driven 3D content generation, but most existing approaches still treat scene generation and user interaction as separate processes, limiting the adaptability and immersive potential of interactive multimedia systems. This paper presents a unified framework that closes the loop between language-driven 3D scene generation and immersive user interaction. Given natural language instructions, the system first constructs structured scene representations using LLMs, and then optimizes spatial layouts via reinforcement learning under geometric and semantic constraints. The generated environments are deployed in a virtual reality setting to facilitate HRI-in-the-loop, where user interactions provide continuous feedback to align generated content with human perception and usability. By tightly coupling generation and interaction, the proposed framework enables more responsive, adaptive, and realistic multimedia experiences. Experiments on the ALFRED benchmark demonstrate state-of-the-art performance in task-based scene generation. Furthermore, qualitative results and user studies show consistent improvements in immersion, interaction quality, and task efficiency, highlighting the importance of closed-loop integration of generation and interaction for next-generation multimedia systems.
Our project page can be found at \url{https://proj-showcase.github.io/h3ds/}. 


%

\IEEEpeerreviewmaketitle
   
\section{Introduction}
\IEEEPARstart{R}ecent advances in LLMs have significantly improved the ability to generate complex 3D environments from natural language, opening new possibilities for content creation in immersive multimedia systems \cite{lu2025vla, liu2026towards, Shao2025DeepSeekMathV2TS, Anh2025, liuhaptic2026}. At the same time, platforms such as virtual reality (VR) are transforming how users experience and interact with digital environments through rich multimodal signals, including visual, spatial, and haptic feedback \cite{Tang2025, LiVR2025, delatorre2024, zdel2025ExploringCA, RAGVR}. Despite these advances, current approaches largely treat content generation and user interaction as separate processes, resulting in static environments that lack adaptability and responsiveness to user behavior. This separation fundamentally limits the potential of next-generation multimedia systems, where seamless integration of generation and interaction is essential for achieving realistic and engaging experiences. 

\begin{figure}[ht!]
\begin{center}
   \includegraphics[width=1\linewidth]{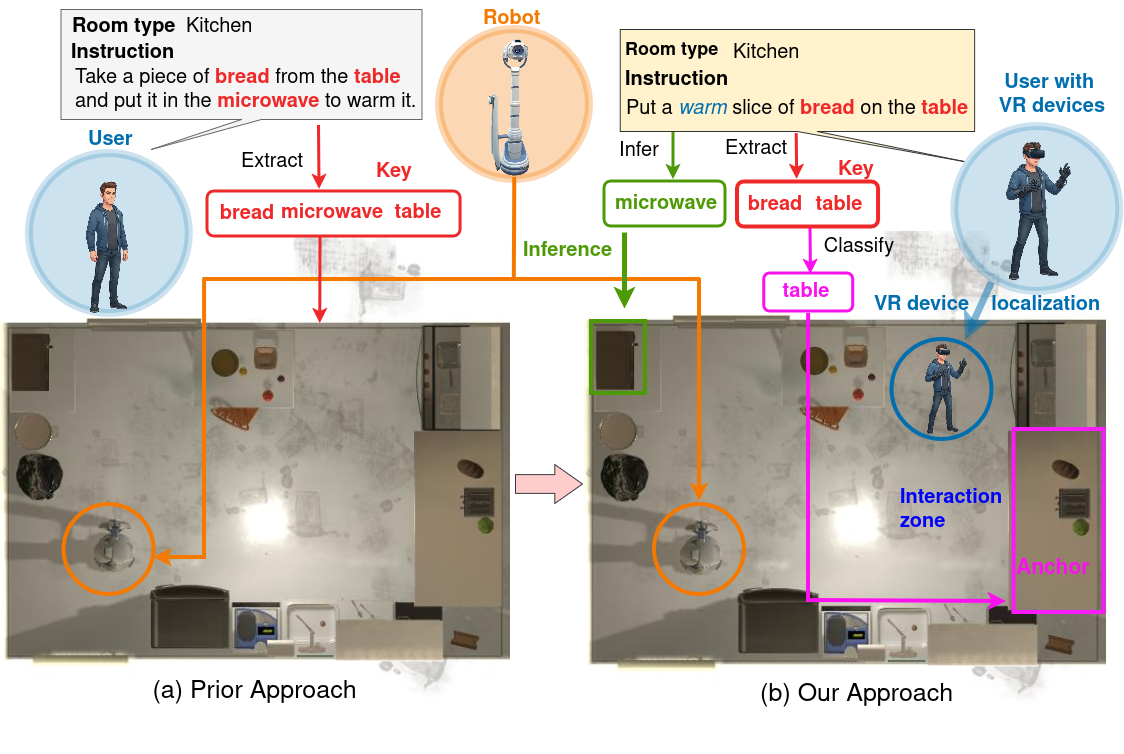}
\end{center}
   \caption{\textbf{Illustration of Our Approach for Language-Driven Scene Generation and Interaction.} (a) Existing methods typically extract explicit "key objects" (e.g., bread, microwave, table) directly mentioned in the instruction to populate the environment. This often results in a functional but static layout that does not account for the implicit requirements of the task or the presence of a human user. (b) Beyond simple extraction, our approach infers necessary auxiliary objects (e.g., identifying that a "warm slice of bread" implies the presence of a microwave) and classifies an anchor object to define the interaction zone. Critically, our model determines an appropriate human position within the 3D scene, facilitating a shared workspace that allows for realistic, immersive interaction between the user and the robot.}
\label{fig:inro_problem}

\end{figure}

For instance, most prior work focuses either on generating static 3D environments from textual descriptions \cite{lin2024instructscene, feng2024layoutgpt, fu2024anyhome, Yang_2024_CVPR, Wang2026LogicEnvGenTD} or on enabling interaction within manually designed environments \cite{rfu2023, Liu2024, wu2021communicative, humanthor}. As a result, there is a fundamental disconnect between content generation and user interaction, limiting scalability, adaptability, and user engagement. In particular, current systems lack a unified mechanism linking language-driven scene generation to real-time multimodal perception and interaction.

To address this limitation, we propose a unified framework that closes the loop between language-driven 3D scene generation and immersive user interaction. Given natural-language instructions, the proposed system first constructs structured scene representations using an LLM and then refines spatial layouts via reinforcement learning under both geometric and semantic constraints. The generated environments are deployed in a VR setting, where users interact with objects through visual and haptic feedback while robotic agents perform task-oriented actions in the same space. This shared environment establishes a closed-loop interaction in which user behavior continuously informs and refines the generated content. By integrating content generation, user perception, and interaction within a single multimodal pipeline, the proposed framework enables more adaptive, responsive, and immersive multimedia experiences.

A key contribution of this work is the explicit modeling of an HRI-in-the-loop paradigm that bridges scene generation with user perception. By deploying the generated 3D environments in a virtual reality setting, the system integrates human users and robotic agents within the same environment. Users interact with objects through multimodal feedback—such as temperature variations and object vibrations—using head-mounted displays and haptic gloves, while robots execute complex, task-oriented actions. This closed-loop mechanism ensures that the interaction loop continuously adapts the scene, enabling seamless and responsive human–robot collaboration.

Building on this design, the proposed framework not only automates the generation of 3D indoor environments from natural language but also integrates multimodal interaction to enhance usability and realism. By combining LLM-based scene understanding, reinforcement learning-based spatial optimization, and VR-based interaction, the system effectively bridges the gap between generative models and interactive multimedia systems. Unlike prior work that treats scene generation and user interaction as separate processes, this work advances multimedia systems by integrating content generation, user perception, and interaction into a unified multimodal loop. This unified design enables generated environments to dynamically adapt to user feedback, resulting in more immersive, responsive, and interactive multimedia experiences.

Extensive experiments on the ALFRED benchmark \cite{ALFRED20} demonstrate state-of-the-art performance in task-based scene generation. Furthermore, qualitative results and user-level evaluations show consistent improvements in immersion, interaction quality, and task efficiency, highlighting the practical benefits of the proposed approach. These findings suggest that tightly coupling generation, perception, and interaction is a promising direction for next-generation multimedia systems. 

The key contributions of this work are summarized as follows:
\begin{itemize}
    \item We introduce a unified framework that closes the loop between language-driven 3D scene generation and immersive user interaction, addressing the fundamental disconnect between content creation and user experience in multimedia systems.
    \item  We propose a language-guided scene representation and reinforcement learning-based spatial optimization strategy that jointly models semantic consistency, geometric feasibility, and interaction readiness, enabling the generation of coherent and usable 3D environments.
    \item We establish an HRI-in-the-loop paradigm within an immersive virtual environment, facilitating seamless collaboration between users and robotic agents while creating a closed-loop system that dynamically adapts to user interactions. 
    \item We validate the proposed framework through extensive experiments on benchmark datasets and user studies, demonstrating state-of-the-art performance in scene generation as well as consistent improvements in immersion, interaction quality, and task efficiency.
\end{itemize}

\section{Related Work}
\subsection{Multimodal Scene Generation with LLMs}
Recent advances in LLMs have enabled significant progress in multimodal content generation, particularly in bridging language and visual representations. In the context of 3D scene synthesis, several works have explored the use of LLMs to generate structured representations from textual descriptions. For instance, InstructScene \cite{lin2024instructscene} and LayoutGPT \cite{feng2024layoutgpt} leverage language-driven priors to construct semantic layouts, while approaches such as AnyHome \cite{fu2024anyhome} utilize prompt-based generation to design indoor environments. However, these methods primarily focus on visual realism and semantic coherence. In contrast, LogicEnvGen \cite{Wang2026LogicEnvGenTD} employs LLMs to generate environments with an emphasis on logical diversity from a testing perspective.

These methods demonstrate the effectiveness of LLMs in transforming natural language into structured scene representations. However, most existing approaches focus on static scene generation and lack mechanisms to support interactive and multimodal experiences. In particular, they do not explicitly consider how generated scenes are perceived and interacted with by users in immersive environments. 
\subsection{LLM-Driven Interactive VR and Multimedia Systems}
With the rapid development of immersive multimedia technologies, LLMs have also been integrated into VR systems to enhance user interaction \cite{Tang2025, LiVR2025, delatorre2024, zdel2025ExploringCA, RAGVR}. Prior work has shown that LLMs can facilitate instruction understanding, context-aware dialogue, and interactive problem solving in VR environments. For example, LLM-based assistants \cite{LiVR2025} have been used in VR scenarios such as escape rooms and question-answering systems, enabling more natural human–AI interaction. In addition, LLMs have been applied to procedural content generation (PCG) and narrative design in multimedia applications, such as generating game quests and interactive storylines \cite{Susanna2024}. These approaches highlight the potential of LLMs to enrich user experience through language-driven interaction.

However, existing systems primarily focus on either interaction or content generation, and rarely unify both aspects within a single framework. In particular, they do not establish a closed-loop interaction between generated content and user perception across multiple modalities, such as visual and haptic signals.
\subsection{Embodied AI and Human–Robot Interaction in Virtual Environments} Embodied AI has emerged as an important paradigm for enabling agents to interact with complex environments. Several works have explored the integration of LLMs with embodied systems for tasks such as navigation, object manipulation, and environment understanding. Frameworks such as Holodeck \cite{Yang_2024_CVPR} and InfiniteWorld \cite{ren2024infiniteworld} enable the generation of interactive environments, while extensions such as DivScene \cite{wang2024divscenebenchmarkinglvlmsobject} and ARCHITECT \cite{wang2024architect} improve scalability and generalization. In parallel, virtual simulation platforms such as RFUniverse \cite{rfu2023}, CollabSphere \cite{Liu2024}, GesTHOR \cite{wu2021communicative} and HumanTHOR \cite{humanthor} have been developed to support human–robot collaboration (HRC) tasks in VR environments. These systems enable users to interact with virtual scenes while coordinating with robotic agents.

Despite these advances, most existing approaches rely on manually designed environments or focus solely on agent-centric interaction. As a result, they lack scalability and fail to fully exploit multimodal interaction mechanisms that integrate user perception, environment generation, and agent behavior.

\subsection{RL for Multimodal Spatial Reasoning}RL has been widely used for sequential decision-making in complex environments, including robotics \cite{SHIN2020113064, li2025hybrid, wu2021communicative, NGUYEN2023110785}, computer vision \cite{wu2025reinforcement}, and language-based tasks \cite{Ouyang2022TrainingLM,liu2026exploratory}. Recent research has explored combining RL with LLMs and vision–language models (VLMs), leading to emerging paradigms such as vision-language-action (VLA) systems \cite{lu2025vla, liu2026towards}. These approaches demonstrate strong potential for multimodal reasoning and decision-making. In 3D scene generation, RL has been used to optimize object placement under geometric constraints \cite{di2022hierarchical}, and layout constraints \cite{Ran2025DirectNL, yang2025optiscene}.

However, prior work typically treats RL as a standalone optimization module and does not fully integrate it into a multimodal interaction pipeline. In particular, the coupling between language-driven scene generation, spatial reasoning, and user interaction remains underexplored. Although existing research has made significant progress in language-driven scene generation, immersive VR interaction, and embodied AI, these directions have largely evolved independently. This fragmentation creates a critical gap between generative content creation and interactive multimedia systems.

\section{Method}
\begin{figure}[ht!]
\begin{center}
   \includegraphics[width=1.05\linewidth]{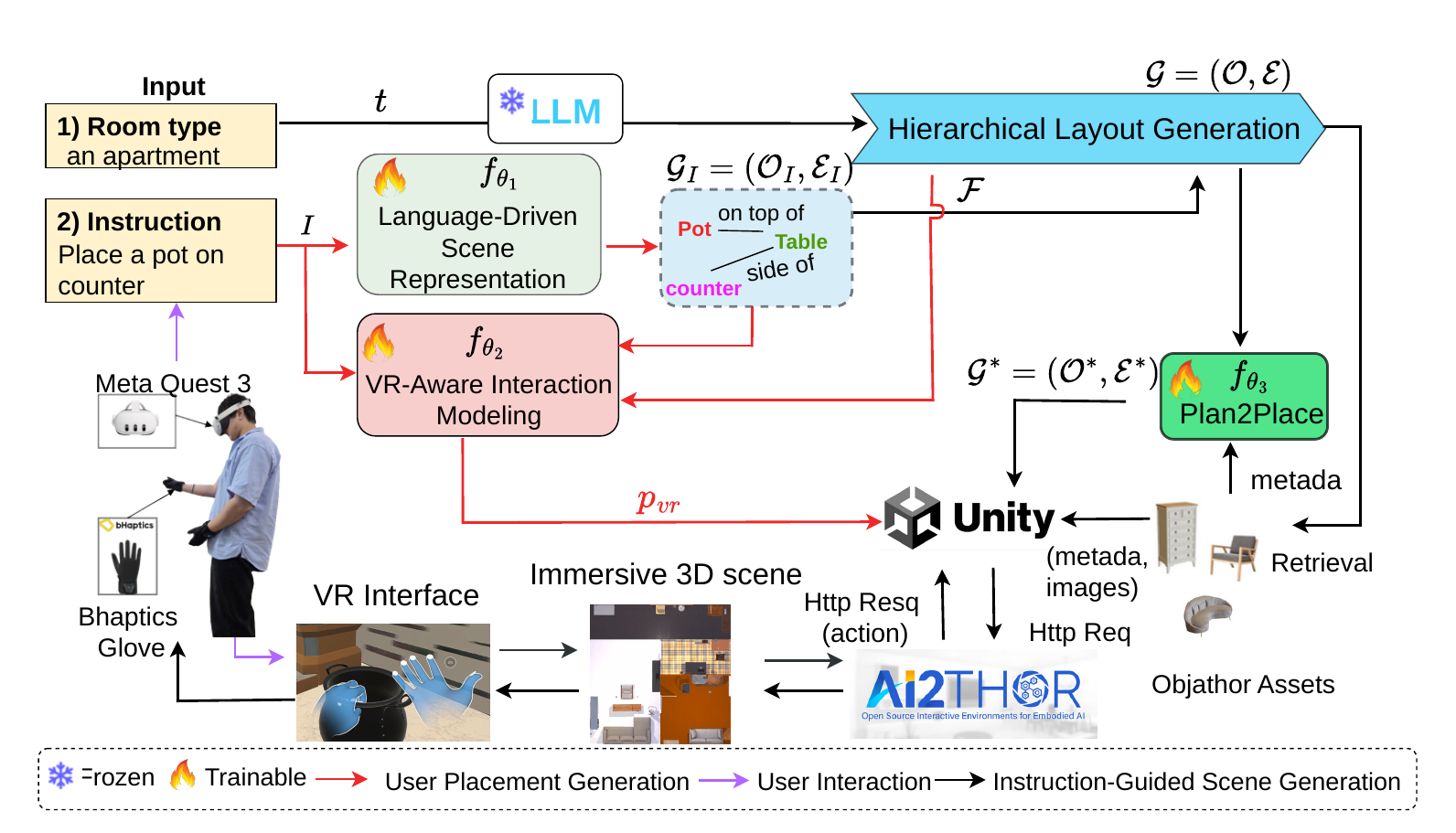}
\end{center}
   \caption{\textbf{System Architecture of Our Multimodal Interaction Framework.} This framework translates high-level linguistic instructions into fully rendered 3D environments. (a) \textbf{Language-Driven Scene Representation}: An instruction ($I$) is processed to create a symbolic scene graph ($\mathcal{G}_I$) defining objects and spatial relationships. (b) \textbf{Hierarchical Layout Generation}: A frozen LLM establishes the global structure (floorplan, walls, ceilings), while VR-Aware Interaction Modeling ($f_{\theta_2}$) determines user perspective ($p_{vr}$). (c) \textbf{Refinement and Rendering}: A final LLM stage synthesizes the layout ($\mathcal{G}$), and the Plan2Place module ($f_{\theta_3}$) handles object placement using Objathor assets. (d) \textbf{Unity+AI2-THOR Integration}: The configuration is sent via HTTP to AI2-THOR and Unity for 3D rendering, supporting real-time interaction via VR hardware.}
\label{fig:overview}
\end{figure}

\subsection{Overview of the Multimodal Interaction Framework}
As illustrated in Fig. \ref{fig:overview}, we propose a unified multimodal framework that integrates language-driven scene generation, spatial reasoning, and immersive interaction within a closed-loop system. Given a natural language instruction, the system first extracts a structured scene representation using an LLM. This representation is then refined through RL-based spatial optimization to construct a coherent 3D environment. The generated scene is rendered in the VR environment, where users interact with objects through visual and haptic feedback. The interaction outcomes are fed back into the system, forming a closed-loop pipeline that connects generation, perception, and interaction. This design enables seamless integration of multiple modalities, including language, vision, and haptic signals, thereby supporting both content generation and real-time interaction. Beyond system integration, this design advances multimedia systems by explicitly unifying content generation, user perception, and interaction within a closed-loop multimodal framework. This integration allows the system to continuously align generated content with human perception and interaction, which is a key requirement for immersive multimedia applications.

\textit{Problem Formulation}:
Given a room type $t$ (e.g. a living room, or an apartment), and an instruction $I$, describing the household task the robot must perform, our objective framework is to automatically generate a 3D immersive environment. This environment supports collaborative tasks between human and robot by integrating  haptic glove and HMD devices.

We use a scene graph $\mathcal{G} = (\mathcal{O}, \mathcal{E})$ to represent the spatial relationships between objects in 3D scene $\mathcal{S}$. Each object $o_i \in \mathcal{O}$ is described by a set of properties $\{\alpha_i, d_i, s_i\}$, where $\alpha_i$ is the object name, $d_i$ is the material description, and $s_i$ is the 3D size. Edges $\mathcal{E}$ capture the adjacency and spatial relationships between pairs of objects. The scene graph is defined as $G = G_t \cup G_I$, where $G_t$ is the scene graph generated by LLM based on the conditional scene type $t$, and $G_I$ is the scene graph generated by  $f_{\theta_1}(G_I|I)$ based on the conditional instruction $I$. Additionally, a function $f_{\theta_2}(p_{vr} \mid \mathcal{G}_I)$ predicts the position of the VR device $p_{vr}$ within the refined room $\mathcal{F^*}$, ensuring consistency with the object set $\mathcal{O}_I$ derived from the room list $\mathcal{F}$ generated by the LLM. Finally, we optimize object placement in $\mathcal{G}$ using the Plan2Place module $f_{\theta_3}(\mathcal{G}^* \mid \mathcal{G})$ resulting in an optimized arrangement of objects. We optimize the weighted parameters by learning $\{\theta_1, \theta_2, \theta_3\}$.  

\subsection{Language-Driven Scene Representation}

To bridge natural language and structured 3D environments, we introduce a language-driven scene representation based on LLMs with Low-Rank Adaptation (LoRA) \cite{hu2021lora}. Given an instruction $I$, an LLM extracts a structured scene graph that encodes both object semantics and spatial relationships. The module learns a mapping: $f_{\theta_1}(G_I|I) \longrightarrow G_I = (\mathcal{O}_I, \mathcal{E}_I)$. 
This representation includes key objects explicitly mentioned in the instruction as well as inferred objects derived from contextual priors. By modeling both explicit and implicit elements, the proposed representation provides a robust interface between high-level language understanding and low-level spatial reasoning. Unlike conventional approaches that directly generate layouts from text, our method explicitly separates semantic understanding from spatial optimization. This design improves generalization and enables flexible integration with downstream modules.  
\subsection{Hierarchical Layout Generation}
\textbf{Floorplan Generation}. 
Similar to \cite{Yang_2024_CVPR}, LLMs receive a tailored prompt to create a floorplan where each room is represented as a rectangle defined by corner coordinates, wall structures, room placement, and connectivity relationships.
Furthermore, the LLM also selects materials for floors and walls to support accurate object retrieval.

\textbf{Doorway and Window Module}.
For each room in the house, the LLMs generate doorway and window properties using separate prompts. These properties include size, height, quantity, style, and more, with styles matched to 40 door types and 21 window types.  

\textbf{Asset Retrieval}. 
The retrieval function used to select objects should be included in the layout. This function computes the visual encoding by the CLIP encoder \cite{pmlr-v139-radford21a} and textual encoding by Sentence Transformers \cite{Reimers2019SentenceBERTSE} similarity and dimensions to ensure the selected assets are suitable for the designed layout.  
\begin{figure*}[ht!]
\begin{center}
   \includegraphics[width=0.99\linewidth]{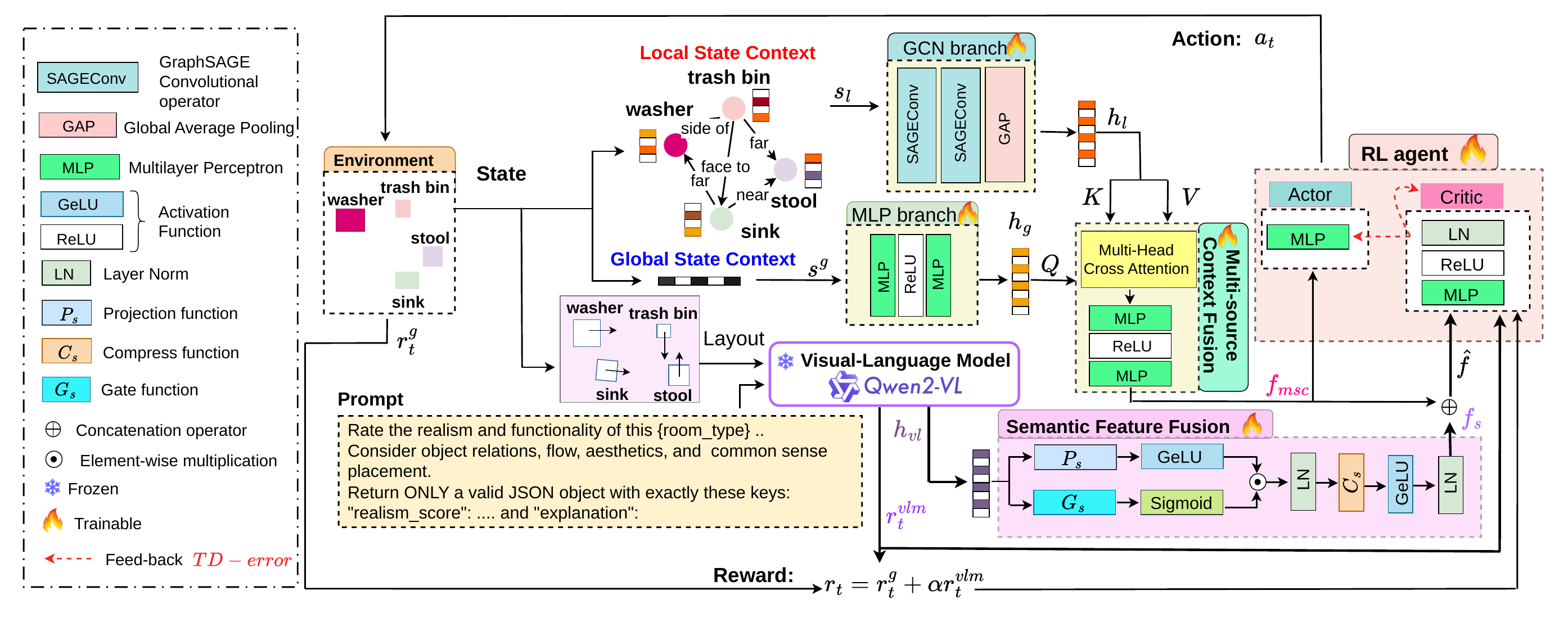}
\end{center}
   \caption{\textbf{Architectural Overview of the Plan2Place Framework.}The pipeline integrates three primary modules to optimize object placement in simulated environments. (a) \textbf{Multi-Source Context Fusion}: this module processes heterogeneous inputs, including Local State Context via a GCN branch using SAGEConv layers and Global State Context via an MLP branch. These features are integrated through a Multi-Head Cross-Attention mechanism to capture complex spatial relationships. (b) \textbf{Semantic Feature Fusion}: High-level semantic evaluation is provided by a frozen Visual-Language Model (Qwen2-VL), which generates a realism score and explanation based on the layout. These semantic features ($h_{vl}$) are refined through a learnable gating mechanism ($G_s$) and projection ($P_s$) to produce a robust multimodal representation ($f_s$). (c) \textbf{RL Agent}: An Actor-Critic architecture utilizes the fused features to determine the optimal action ($a_t$). The reward function ($r_t$) is formulated as a combination of environmental feedback ($r_t^g$) and VLM-based semantic evaluation ($r_t^{vlm}$), scaled by a hyperparameter $\alpha$.}
\label{fig:plan2place_system}
\end{figure*}

\subsection{VR-Aware Interaction Modeling}

To enable immersive interaction, we introduce a VR-aware modeling module that predicts optimal interaction viewpoints within the generated environment. Given the scene graph $\mathcal{G}_I = (\mathcal{O}_I, \mathcal{E}_I)$, the model estimates suitable positions for VR devices by classifying object roles based on their relevance to user interaction. We employ Bidirectional Encoder Representations from Transformers (BERT) \cite{Devlin2019BERTPO} to capture contextual relationships between objects and instructions, allowing the system to identify anchor objects and interaction regions. This module ensures that the generated environment is not only structurally coherent but also interaction-ready. By explicitly incorporating user perception into the generation pipeline, the proposed approach bridges the gap between scene synthesis and immersive multimedia interaction.

Given the orginal dataset $D = \{(x'_i, Y_i)\}_{i=1}^{N}$ where $x'_i$ is the instruction and $Y_i = \{ y_k, O'_{ik}\}_{k=1}^{K}$ is a mapping of object category $y_k \in \{ \text{anchor object, key object, inference object} \}$ to a set of object names (entities) $O'_{ik}$. $K$ denotes the number of object types, including anchor, key, and inference.
We learn a function $f_{\theta_2}$ that predicts the probability of each object category for a given entity, conditioned on the instruction. Specifically, we jointly encode the entities in $\mathcal{O}_I$ and the instruction $I$ as a unified textual sequence, allowing the model to capture their contextual interactions. A BERT encoder is then used to model bidirectional dependencies, enabling the representation of each entity to be informed by both the instruction semantics and its relationships with other entities. This design helps the model better handle rare or unseen objects by leveraging long-range contextual cues within the instruction.  
The input text is formed as:
\begin{equation}
\begin{aligned}
\mathbf{w} = \mathcal{BERT}_{tokenizer}(x'_i \quad \text{[SEP]} \quad o')
\end{aligned}
\end{equation}
where [SEP] is the BERT separator token. The tokenizer outputs $\mathbf{w}$, including token indices and a binary mask for padding $\in \{0, 1\}^L$ where $L = 128$ is the maximum sequence length.

Next, the extracted texture  features are fed into an object classification module comprising two linear layers, which perform nonlinear dimensionality reduction to distill the salient characteristics of the object, resulting in the final latent representation $\hat{\mathbf{w}}$

\textit{Training Objective}: A multi-label classifier $f_{\theta_2}$ is used for the model output as follows:
\begin{equation}
\begin{aligned}
\hat{y}_{i,e} = f_{\theta_2}(\hat{\mathbf{w}}) \in [0, 1]^K
\end{aligned}
\end{equation}
and is typically trained with binary cross-entropy loss:
\begin{equation}
\begin{aligned}
\mathcal{L}_{\theta_2} = -\displaystyle \frac{1}{\mid \mathcal{D} \mid} \sum_{(x',o',y) \in \mathcal{D}} \sum_{k=1}^K [y_k\text{log} \hat{y}_k - (1 - y_k)\text{log}(1 - \hat{y}_k)]
\end{aligned}
\end{equation}

A few-shot prompt approach is adopted to identify the room type $\mathcal{F}^*$ based on the given set of objects $\mathcal{O}_t$. This approach enables our model to flexibly infer various room types generated by LLM without being restricted to predefined categories in the dataset. See Supplementary A-F 
for prompt details.

\subsection{Plan2Place: RL-based Object Placement Optimization}
Existing studies have typically employed reinforcement learning to optimize object placement using geometric information \cite{di2022hierarchical} or have adopted Reinforcement Learning from Human Feedback (RLHF) frameworks such as Direct Preference Optimization (DPO) \cite{Ran2025DirectNL, yang2025optiscene}. However, these approaches either focus solely on geometric constraints between objects, neglecting the semantic relationships in the generated layouts, or rely on large-scale, high-quality preference datasets (e.g., chosen/rejected pairs), which are expensive and labor-intensive to construct.

To address these challenges, we propose Plan2Place ($f_{\theta_3}$), an online RL-based framework for object placement optimization under multimodal constraints as illustrated in Fig.~\ref{fig:plan2place_system}, where the task is formulated as a sequential decision-making process. The agent determines object placements to satisfy both geometric constraints (e.g., collision avoidance and spatial feasibility) and semantic constraints (e.g., relative positioning and functional relationships), ultimately producing realistic, interaction-ready layouts.

The environment state encodes both global layout context and local object relationships through graph-based representations. To capture geometric information from the current state, we introduce a Multi-Source Context Fusion (MSCF) module. In parallel, a Semantic Feature Fusion (SFF) module extracts semantic features from the layout using a pre-trained VLM. The fused features are then passed to the critic network to capture multimodal consistency between object configurations and semantic expectations, thereby guiding the agent toward improved placement decisions.

To further promote realistic and semantically consistent layouts, we design a reward function that accounts for geometric validity, relational consistency, and interaction affordances. Compared to heuristic or rule-based methods, Plan2Place provides a more adaptive and scalable solution, making it well-suited for complex and diverse environments.

\subsubsection{Object Placement Environment}
Similar to \cite{Yang_2024_CVPR, feng2024layoutgpt}, Plan2Place receives LLM-derived constraints, the room size, and object information (i.e., names and bounding boxes) as input.

\textbf{State}: We represent both global and local state context describing the environment and spatial relationships. 

The global state, denoted as $\mathbf{s}_g \in \mathbb{R}^{1 \times 5}$, encodes high-level information about the scene, including the number of global constraints (e.g., edge or center placement), the size ratio of the next object relative to the room dimensions, and the proportion of remaining objects to be placed. A two-layer MLP with ReLU activation produces the global feature representation:
\begin{equation}
\mathbf{h}^{g} = \mathbf{W}_2(\text{ReLU}(\mathbf{W}_1\mathbf{s}_g + \mathbf{b}_1) + \mathbf{b}_2),
\end{equation}
where $\mathbf{W}_1 \in \mathbb{R}^{h_d \times 5}$ and $\mathbf{W}_2 \in \mathbb{R}^{h_d \times h_d}$, with $h_d = 128$ in our experiments. 

The local state ($\mathbf{s}_l$) is represented as a spatial relational graph, where each node corresponds to an object and edges encode pairwise spatial relationships. The graph is parameterized by a node feature matrix $\mathcal{X} \in \mathbb{R}^{M \times 518}$ and an adjacency matrix $\mathcal{A} \in \mathbb{R}^{M \times M}$, where $M$ is the number of objects. Each node feature $\mathbf{x}_i$ integrates both geometric and semantic information, including normalized size ratios, spatial position, orientation, and semantic embeddings derived from object categories using CLIP, while edge features capture relative spatial  configurations, including distances, angles, alignment scores, and constraint-based relation vectors.
To encode the local structural context, we employ a two-layer GraphSAGE encoder followed by a global average pooling operator to obtain a compact representation:
\begin{equation}
\mathbf{h}^l = \mathrm{GAP} \left( \mathrm{GraphSAGE}(\mathcal{X}, \mathcal{A}) \right),
\end{equation}
where $\mathbf{h}^l \in \mathbb{R}^{h_d}$ denotes the aggregated local feature vector. This process allows the model to capture the spatial relationships among neighboring features.

\textit{Multi-Source Context Fusion.}
MSCF learns a unified representation by modeling cross-scale interactions between global layout context and local spatial relationships via multi-head cross-attention. The context-aware feature $\mathbf{h}^{\text{fused}} \in \mathbb{R}^{h_d}$ is computed from $\mathbf{h}^g$ and $\mathbf{h}^l$, encoding both global structural information and fine-grained spatial dependencies. A lightweight feed-forward network further refines $\mathbf{h}^{\text{fused}}$ to yield the final geometric feature $\mathbf{f}_{\text{msc}}$, which is used by the policy and critic.

\textit{Semantic Feature Fusion.}
To extract semantic features from the layout, SFF leverages multimodal representations from a pre-trained VLM. Specifically, the layout representation, comprising object positions, bounding boxes, and orientations, together with a textual prompt, is fed into the VLM to evaluate the quality of the input layout. This process obtains a semantic feature $\mathbf{h}_{vl} \in \mathbb{R}^{2048}$ and a semantic score $r_t^{vlm}$. 

A projection and gating mechanism maps $\mathbf{h}_{vl}$ into a task-adaptive latent space and performs feature-wise modulation, followed by normalization and compression to obtain a compact semantic representation $\mathbf{f}_s \in \mathbb{R}^{h_d/2}$. This representation preserves salient semantic information while enhancing alignment with geometric features for subsequent fusion.

\textbf{Action}: At each time step $t$, the agent selects an action $a_t$ that determines the spatial placement of the current object within the scene. The action is parameterized by the object’s position $\mathbf{p}_t \in \mathbb{R}^2$ and orientation $\theta_t$ on the floor plane, with $\mathbf{p}_t$ constrained to lie within the valid room region $\mathcal{R}$, which reduces the search space while preserving spatial diversity.

\textbf{Reward}: We formulate the reward as the negative of an energy function that measures the degree of constraint violation in the generated layout.

The total energy combines several components:
\begin{equation}
E_{\text{total}} = \lambda_1 \hat{E}_{rel} + \lambda_2 \hat{E}_{collision} + \lambda_3 \hat{E}_{oob} + \lambda_4 \hat{E}_{nav} + \lambda_5 \hat{E}_{aff},
\end{equation}
where $\hat{E}_{rel}$ denotes relational energy derived from LLM-based constraints and human-designed priors (e.g., 3D-FRONT, ProcTHOR), $\hat{E}_{collision}$ measures object overlap, $\hat{E}_{oob}$ penalizes out-of-bound placements, $\hat{E}_{nav}$ reflects navigation feasibility, and $\hat{E}_{aff}$ encodes interaction affordances via clearance regions in front of functional objects.  
The weighting coefficients $\lambda_{1\text{--}5}$ are empirically set to 4, 1.5, 0.35, 1.5, and 1.5, respectively. Additional details on the reward energy functions are provided in Suppl. A-1 

The RL reward is then defined as $R_t = -E_{\text{total}}$, encouraging physically valid, navigable, and semantically coherent layouts.

\subsubsection{Policy Network}
Among many alternatives, we adopt an actor–critic network \cite{SHIN2020113064} to optimize the action selection process, as it balances flexibility and stability for real-world layout generation. The actor network learns a policy to select actions, while the critic network estimates the expected return based on the multimodal state representation.

\textit{Policy Loss}: The policy is optimized to favor actions with higher advantage:
\begin{equation}
\mathcal{L}_{policy} = \frac{-1}{T} \sum_{t=1}^T \log \pi(a_t \mid s_t) \, A_t,
\end{equation}
where $A_t$ is the advantage and $s_t$ corresponds to the fused geometric feature $\mathbf{f}_{msc}$.

\textit{Value Loss}: The critic learns to estimate the expected return $V_t$ from the combined representation $\hat{f} = \mathbf{f}_{msc} \bigoplus \mathbf{f}_s$:
\begin{equation}
\mathcal{L}_{value} = \frac{1}{T} \sum_{t=1}^T (V_t - R_t)^2.
\end{equation}
where $R_t$ is the target return at time step $t$.

\textit{Auxiliary Loss}: To incorporate semantic guidance, we align the critic’s predictions with the VLM-derived semantic scores $r_t^{vlm}$:
\begin{equation}
\mathcal{L}_{aux} = \sum_{t=1}^T \mathbb{I}_t (V_t - r_t^{vlm})^2,
\end{equation}
where $\mathbb{I}_t$ indicates the availability of a semantic score.

\textit{Entropy}: An entropy bonus maintains sufficient exploration.
\begin{equation}
\mathcal{H} = \frac{-1}{T} \sum_{t=1}^T \sum_{a} \pi(a \mid s_t) \log \pi(a \mid s_t).
\end{equation}

The total loss is:
\begin{equation}
\mathcal{L} = \lambda^{p}\mathcal{L}_{policy} + \lambda^{v}\mathcal{L}_{value} + \lambda^{aux}\mathcal{L}_{aux} - \lambda^{e}\mathcal{H},
\end{equation}
where $\lambda^{p}$, $\lambda^{v}$, $\lambda^{aux}$, and $\lambda^{e}$ are empirically set to 1, 0.5, 0.1, and 0.01, respectively.

\subsection{Multimodal Interaction Loop Integration} 
The generated 3D environments are rendered in a VR setting, where users interact with objects through visual and haptic modalities. These interactions provide implicit feedback on the quality and usability of the generated scenes. By incorporating this feedback into the system pipeline, the framework continuously aligns generated content with user perception. Unlike conventional pipelines that decouple generation and interaction, this work advances multimedia systems by integrating content generation, user perception, and interaction into a unified multimodal loop. This closed-loop mechanism enables adaptive scene generation that responds to user behavior, thereby enhancing both realism and usability in immersive environments.

Our system, as shown in Fig. \ref{fig:haptic_system}, employs several APIs for interacting with 3D immersive environments generated by LLMs, as described below:
\begin{figure}[ht!]
\begin{center}
   \includegraphics[width=0.8\linewidth]{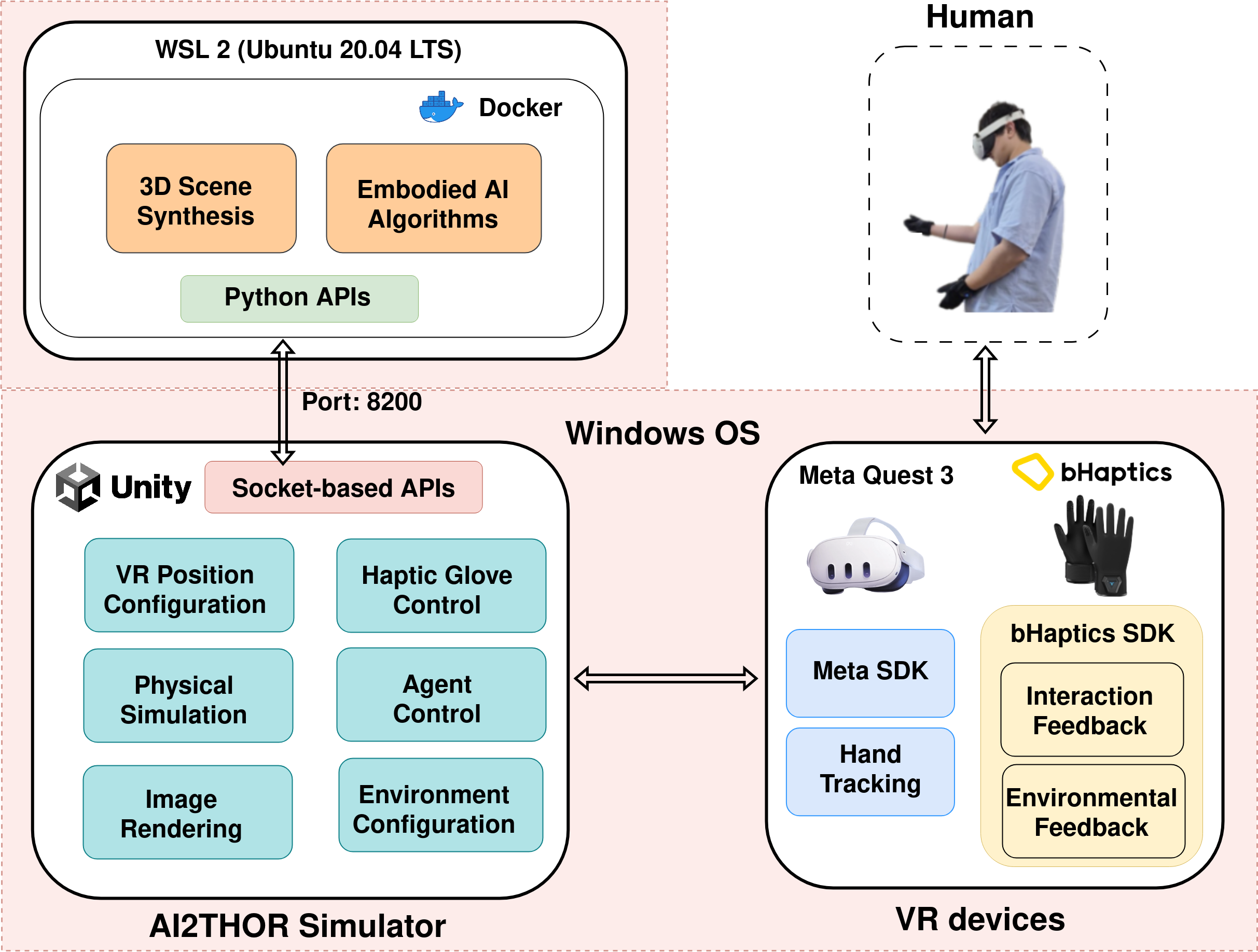}
\end{center}
   \caption{\textbf{Three-Tiered System Architecture for the VR-Integrated Embodied AI Framework.} This cross-platform infrastructure balances high-load computational reasoning with low-latency interaction: (a) \textbf{Creation and Inference Layer (WSL 2/Ubuntu)}: A Dockerized environment handling core algorithmic workloads like 3D Scene Synthesis and Embodied AI, communicating via Python-based Socket APIs (Port 8200). (b) \textbf{Execution Layer (Windows/Unity)}: Acts as the bridge between AI models and hardware, utilizing AI2-THOR for physical simulation, rendering, environment configuration, and real-time processing of agent and haptic commands. (c) \textbf{VR Device Layer}: Enables human-in-the-loop interaction, leveraging the Meta SDK for hand tracking and visual output, with the bHaptics SDK providing granular environmental and interaction feedback.}
\label{fig:haptic_system}
\end{figure}

\textit{VR Position Configuration}: 
We introduce a VR position configuration module that automatically determines the placement of VR devices within a generated VR environment, using the object ID predicted in the previous stage. 

To determine feasible placements, the module adopts a sampling-based strategy that generates candidate positions around the target object at a fixed distance. Specifically, this module first computes the object's center from its bounding box, then performs multiple random sampling trials by selecting directions on the horizontal plane and projecting them into 3D space to generate candidate positions.

Each candidate is temporarily assigned to the VR device transform and evaluated through two validity checks: interaction with the target object and collision avoidance with environmental obstacles (e.g., walls). Candidates that satisfy both conditions are considered valid. The module then computes their distances to the room center and stores them along with their corresponding directions, enabling subsequent ranking and selection of the most suitable placement.

\textit{Haptic Glove Control}: The bHaptics SDK is utilized to integrate and manage the haptic glove, which provides real-time vibration feedback to enhance user immersion. Our system provides multimodal feedback to simulate interactions in the virtual environment, using vibrations for touch sensations and visual cues (e.g., color changes) to represent changes in object or environmental states, such as temperature.

Specifically, colliders are attached to each fingertip to detect direct hand–object interactions. When a fingertip touches or grasps an object, the Interactable Unity Event Wrapper system identifies the interaction state (i.e., hover, unhover, select, or unselect). The detected object name is then used as an input key for subsequent processing.

For each object, the system queries its physical attributes, including pickupable, heat source, cold source, and mass to classify the object as hot, cold, light, or normal. Based on this classification, the system generates corresponding visual (hand color) and haptic (vibration) feedback.

For example, in temperature-related scenarios, users perceive changes through vibration feedback, while the glove’s color transitions from black to blue to indicate decreasing temperature. In lighting-related scenarios, the glove’s color gradually shifts from black to white to reflect increasing brightness. This multimodal feedback mechanism allows users to perceive environmental changes more intuitively, resulting in a more realistic and immersive human–robot collaboration experience.

\subsection{Dataset.}
\textit{1) The Task-based Generation Dataset}.
To evaluate our method, we selected four tasks from the ALFRED dataset \cite{ALFRED20} that are well-suited for integration with VR devices. These tasks include “pick up and place (simple),” “look at object in light,” “pick, cool, then place in receptacle,” and “pick, heat, then place in receptacle.” These tasks enhance user experience by allowing users to interact with objects through specific VR devices. The dataset contains 11,306 samples for training, 989 samples for validation, and 500 samples for unseen validation.

We further evaluate our model using these evaluation sets as follows: i) \textit{In-Distribution (ID)}: this set consists of 500 unseen validation scenes sampled from the same distribution as the training data; ii) \textit{Object Shift (OS)}: we generated this split of 500 samples by rephrasing the linguistic templates of instructions in the ID set, while preserving their original semantic meaning. This setup evaluates the model's robustness to variations in instruction phrasing; iii) \textit{Object Shift (OS)}: this set is constructed by replacing objects in the ID set with novel or less frequent categories, testing the model's ability to generalize to unseen object combinations and semantic compositions. In addition, we manually constructed ground-truth scene graphs for instructions across four selected tasks to facilitate targeted analysis. See Suppl. A-A 
for details.

\textit{2) The 3D Scene Datasets}

- \textit{Prior Datasets}.
To facilitate learning of semantic spatial relationships between objects and to ensure aesthetically coherent placements, we filter approximately 3,587 scenes from 3D-FRONT \cite{fu20213d} and ProcTHOR \cite{procthor2022}, focusing on independent room types such as hallways, kitchens, bedrooms, bathrooms, studies/offices, dining rooms, storage rooms, balconies, and living rooms. For simplicity, our current framework is limited to rooms with rectangular or square layouts.

- \textit{The LLM-based Constraint Datasets}.
Building on prior work \cite{Yang_2024_CVPR}, we utilize Llama 3 (8B), Mistral (7B), and Qwen2.5 to generate three constraint datasets for scene layout tasks. Instructions are sourced from the ALFRED dataset and paired with nine representative room categories (hallway, kitchen, dining room, bedroom, laundry room, garage, office, living room, and bathroom) to enable scalable apartment layout generation. After filtering, the LLaMA-based constraint dataset contains 6,100 scene layouts, while the Qwen- and Mistral-based datasets include 770 and 640 scene layouts, respectively.

\subsection{Experiment Setting}
\subsubsection{Configuration Setting}
All experiments, including training and testing, were conducted using a workstation equipped with an AMD PRO 5975WX (32-cores) CPU and an NVIDIA RTX A6000 GPU, running on an Ubuntu 20.04 LTS operating system. 
 
\subsubsection{VR Devices Setting}
To develop and interface with the Meta Quest 3 in a Windows PC environment, we used three software components.

\textit{- Meta XR SDK}: This is the core SDK for our Unity project. It allows the Unity engine to communicate directly with the Quest 3 headset's hardware. 

\textit{- Oculus PC App}: This is an essential program that acts as a bridge connecting the Quest 3 and the PC. 

\textit{- Meta Quest Developer Hub}: This hub allows us to manage various auxiliary functions required for development.

To implement the sophisticated tactile feedback of the haptic gloves, we utilized the following bHaptics software ecosystem.

\textit{- bHaptics SDK (for Unity)}: Utilized to implement and execute haptic feedback within the Unity environment.

\textit{- bHaptics Player}: It maps Unity SDK commands to vibration patterns on Bluetooth haptic gloves.

More details of each experimental setup are provided in Suppl A-C
\subsection{Results}
\begin{table*}[ht!]
    \centering
    \resizebox{2\columnwidth}{!}{
    \begin{tabular}{|*{11}{c|}}  
    \hline
    \multirow{2}{*}{\textbf{Approach}} & \multirow{3}{*}{\textbf{Method}} & \multicolumn{3}{|c|}{ID} & \multicolumn{3}{|c|}{TS} & \multicolumn{3}{|c|}{OS} \\ 
    \cmidrule{3-11}
     & & \textbf{F1} $\uparrow$ & \textbf{iRecall} $\uparrow$ & \textbf{GED} $\downarrow$ & \textbf{F1} $\uparrow$ & \textbf{iRecall} $\uparrow$ & \textbf{GED} $\downarrow$  & \textbf{F1} $\uparrow$ & \textbf{iRecall} $\uparrow$& \textbf{GED} $\downarrow$ \\ 
    \hline 
IM & Vicuna-7B & 81.30 $\pm$ 0.10 & 54.47 $\pm$ 0.22 & 4.09 $\pm$ 0.02 & 81.37 $\pm$ 0.00 & 53.39 $\pm$ 0.00 & 4.00 $\pm$ 0.00 & 80.03 $\pm$ 0.00 & 51.67 $\pm$ 0.00 & 4.4 $\pm$ 0.00\\
     \cline { 2 - 2}& Alpaca & 80.59 $\pm$ 0.08 & 45.36 $\pm$ 0.38 & 4.00 $\pm$ 0.03 & 80.01 $\pm$ 0.10 & 46.92 $\pm$ 0.66 & 4.29 $\pm$ 0.03 & 77.27 $\pm$ 0.07& 43.98 $\pm$ 0.38& 4.64 $\pm$ 0.02\\
    \hline
    Seq2Seq  & T5 & 71.11 $\pm$ 0.00 & 30.38 $\pm$ 0.00 & 5.04 $\pm$ 0.00 & 73.28 $\pm$ 0.00 & 35.16 $\pm$ 0.00 & 4.45 $\pm$ 0.00 & 68.17 $\pm$ 0.00 & 36.99 $\pm$ 0.00 & 4.51 $\pm$ 0.00\\
           \cline { 2 - 2}& Flan-T5 & 78.05 $\pm$ 0.00 & 39.31 $\pm$ 0.00 & 3.81 $\pm$ 0.00 & 79.57 $\pm$ 0.00 & 46.32 $\pm$ 0.00 & 3.54 $\pm$ 0.00 & 76.67 $\pm$ 0.00 & 47.43 $\pm$ 0.00 & 3.48 $\pm$ 0.00\\
    \hline
     CLM & LLaMA 3.1-8B & 82.16 $\pm$ 0.06 & 64.46 $\pm$ 0.36 & 3.88 $\pm$ 0.02 & 81.87 $\pm$ 0.11 & 59.65 $\pm$ 0.39 & 4.08 $\pm$ 0.02& 80.44 $\pm$ 0.08& 58.86 $\pm$ 0.29& 4.27 $\pm$ 0.03\\
    \cline { 2 - 2} & Mistral-7B &  \textbf{84.28 $\pm$ 0.05} & \textbf{68.58 $\pm$ 0.31} & \underline{2.58 $\pm$ 0.02} & \underline{83.97 $\pm$ 0.02} & \underline{62.93 $\pm$ 0.11} & \underline{2.75 $\pm$ 0.01}& \underline{82.72 $\pm$ 0.02} & \underline{63.46 $\pm$ 0.27} & \underline{2.75 $\pm$ 0.01}\\
    \cline { 2 - 2} & Falcon-7B & \underline{84.20 $\pm$ 0.04} & \underline{64.93 $\pm$ 0.11} & \textbf{2.51 $\pm$ 0.00} & \textbf{84.90 $\pm$ 0.01} & \textbf{65.07 $\pm$ 0.09} & \textbf{2.52 $\pm$ 0.00} & \textbf{83.92 $\pm$ 0.02} & \textbf{66.38 $\pm$ 0.12} & \textbf{2.34 $\pm$ 0.00}\\
    \hline
    \end{tabular}
    }
    \caption{Comparison of language-driven scene representation performance from input prompts using different methods across the evaluation sets: In-Distribution (ID), Template Shift (TS), and Object Shift (OS).}
    \label{tab:llm-scene-graph}
\end{table*}

\begin{table}[t]
    \centering
     \resizebox{0.9\columnwidth}{!}{
    \begin{tabular}{|*{8}{c|}}  
    \hline
    \multirow{2}{*}{\textbf{Dataset}} & \multirow{3}{*}{\textbf{Class}} & \multicolumn{3}{|c|}{\textbf{Acc} $\uparrow$}\\ 
    \cmidrule{3-5}
     & & LLM-E2E & LLM+FH & Ours\\ 
    \hline
   ID & Key & 44.82 $\pm$ 0.0013 &  70.33 $\pm$ 0.0004 & \textbf{99.94 $\pm$ 0.0000}\\
   \cline { 2 - 2}
   & Anchor & 70.86 $\pm$ 0.0048 & 67.22 $\pm$ 0.0004 &  \textbf{98.79 $\pm$ 0.0000}\\
    \cline { 2 - 2}
   & Inference & 25.62 $\pm$ 0.0009 &  70.27 $\pm$ 0.0004  &  \textbf{100.0 $\pm$ 0.0000}\\
   \hline
   TS & Key & 37.17 $\pm$ 0.0100 & 74.91 $\pm$ 0.0010 & \textbf{99.87 $\pm$ 0.0000}\\
   \cline { 2 - 2}
   & Anchor & 74.67 $\pm$ 0.0267 & 66.63 $\pm$ 0.0006 & \textbf{92.65$\pm$ 0.0006}\\
   \cline { 2 - 2}
   & Inference & 22.50 $\pm$ 0.0403 & 75.10 $\pm$ 0.0010 & \textbf{99.93 $\pm$ 0.0000}\\
   \hline
   OS & Key & 52.08 $\pm$ 0.0017 & 75.18 $\pm$ 0.0010 & \textbf{99.94 $\pm$ 0.0000}\\
     \cline { 2 - 2}
   & Anchor & 70.10 $\pm$ 0.0033 & 66.14 $\pm$ 0.0005 & \textbf{94.57 $\pm$ 0.0005}\\
    \cline { 2 - 2}
   & Inference & 22.85 $\pm$ 0.0024 & 75.25 $\pm$ 0.0010 & \textbf{99.94 $\pm$ 0.0000}\\
    \hline
    \end{tabular}
    }
    \caption{Comparison of object type prediction performance across three evaluation sets: ID, TS, and OS.}
    \label{tab:llm-pos-haptic}
\end{table}

\subsubsection{Language-Driven Scene Representation}
We evaluate three types of LLM-based approaches: 1) \textit{Causal Language Models (CLM)}, including LLaMA 3.1-8B \cite{Touvron2023LLaMAOA}, Mistral-7B \cite{Jiang2023Mistral7}, and Falcon-7B \cite{Almazrouei2023TheFS}; 2) \textit{Instruction-Tuned Models (ITM)}, including Vicuna-7B and Alpaca-7B \cite{alpaca}; 3) \textit{Sequence-to-Sequence Models (Seq2Seq)}, including The Text-to-Text Transfer Transformer (T5) \cite{2020t5}, and Flan-T5 \cite{Chung2022ScalingIL}.

Performance is evaluated using widely adopted metrics for graph and scene generation tasks: Graph Edit Distance (GED) \cite{wang2021combinatorial, chen2019efficient}, Instruction Recall (iRecall) \cite{lin2024instructscene}, and F1 \cite{jiao2023instruct}. To optimize the learning process of each LLM model, we configured LoRA with $\alpha$ of 43, a dropout rate of 0.05, and a matrix rank ($r$) of 32.
Detailed definitions and implementation are provided in Suppl. A-B 

Results shown in Table \ref{tab:llm-scene-graph} demonstrate that the CLM-based approach achieve higher and more stable performance across the three test sets compared to ITM or Seq2Seq approaches, considering all evaluation metrics. Among them, Falcon-7B fine-tuned delivers the best overall performance. Although it slightly underperforms Mistral in terms of F1 score and iRecall on ID, its GED remains superior on the same dataset. For the remaining test sets, Falcon-7B consistently outperforms nearly all competing methods.

\subsubsection{Position Prediction of VR Devices} 
When a human user wears VR devices and interacts with objects in an immersive 3D scene, we aim to determine the position where the user can observe within the room. To address this problem, we formulate the prediction of VR device positions as a classification task.
We evaluate our method with baselines: 1) an end-to-end LLM-based method (LLM-E2E) is used to directly predict object categories and generate task-based scene graphs; 2) an object classification model that combines a pre-trained LLM encoder for extracting rich textual features with a MLP head to accurately classify object types (LLM+FH).

We employ Falcon-7B due to its strong performance in task-oriented generation, producing both scene graphs and object category predictions. Table \ref{tab:llm-pos-haptic} summarizes the performance of each model across all object types. Additional results evaluated using various metrics can be found in Suppl.A-D 
The results show that our method consistently outperforms both LLM-E2E and LLM+FH approaches, achieving over 90\% accuracy for most object categories. These results are explained by BERT’s self-attention mechanism, which captures bidirectional context and models long-range dependencies between tokens. This enables the model to generalize more effectively, particularly when handling rare or previously unseen objects. 

\subsubsection{Object Placement Optimization}
\label{label:main_plan2place}
\begin{table*}[ht!]
    \centering
    \centering
     \resizebox{1.5\columnwidth}{!}{
    \begin{tabular}{|*{11}{c|}}
    \hline
     \multirow{2}{*}{\textbf{Dataset}} & \multirow{3}{*}{\textbf{Method}} & \multicolumn{2}{|c}{Fidelity} & \multicolumn{3}{|c|}{Plausibility} & 
     \multirow{2}{*}{Avg.} \\
    \cmidrule{3-7}
    &  & \textbf{CNT} $\uparrow$ & 
    \textbf{SR} $\uparrow$ & \textbf{NAV} $\uparrow$ & \textbf{Key\_NAV} $\uparrow$ & \textbf{OOB} $\downarrow$ & PTO $\downarrow$\\ 
    \hline 
    & DFS & 80.51 $\pm$ 0.0014 & 
        \textbf{93.68 $\pm$ 0.0000} & 99.84 $\pm$ 0.0002 & 45.93 $\pm$ 0.0091 & 7.13 $\pm$ 0.0113 & 1.5278 \\
    \cline{2-2}
  LLaMA  & MILP & 67.08 $\pm$ 0.0024 & 
    93.67 $\pm$ 0.0000 & 99.18 $\pm$ 0.0004 & 52.14 $\pm$ 0.0081 & 0.0344 $\pm$ 0.0119 & 0.8573 \\
    \cline{2-2}
 (seen) & Z3  & 41.40 $\pm$ 0.0485 & 
    81.29 $\pm$ 0.0053 & 87.06 $\pm$ 0.0045 & 54.34 $\pm$ 0.0253 & 0.0 & 14.0805 \\
    \cline{2-2}
    & \textbf{Ours} & \textbf{80.68 $\pm$ 0.0011} & 
    \textbf{ 93.68 $\pm$ 0.0000} & \textbf{ 99.93 $\pm$ 0.0003} & \textbf{ 56.79 $\pm$ 0.0053} & \underline{6.6232 $\pm$ 0.0216} & \textbf{0.679} \\
    \hline
    & DFS & 73.83 $\pm$ 0.0011 & 
    84.05 $\pm$ 0.0000 & 99.35 $\pm$ 0.0005 & 33.98 $\pm$ 0.0046 & 5.4659 $\pm$ 0.0732 & 1.3093 \\
    \cline{2-2}
 Qwen & MILP & 50.58 $\pm$ 0.0044 &
    83.86 $\pm$ 0.0013 & 98.37 $\pm$ 0.0011 & \textbf{45.63 $\pm$ 0.0095} & 0.0333 $\pm$ 0.0062 & \textbf{0.7066} \\
    \cline{2-2}
 (unseen)   & Z3  & 15.68 $\pm$ 0.0118  &  
    71.80 $\pm$ 0.0257 & 86.38 $\pm$ 0.0111 & 36.04 $\pm$ 0.0258 & 0.0
    & 6.4757 \\
    \cline{2-2}
     & \textbf{Ours}  & \textbf{78.53 $\pm$ 0.0020} & 
    \textbf{84.16 $\pm$ 0.0000} & \textbf{ 99.67 $\pm$ 0.0007} &  35.65 $\pm$ 0.0086 & 3.6685 $\pm$ 0.0312 & \underline{1.2050} \\
    \hline
    & DFS & 60.79 $\pm$ 0.0016 & 
    \textbf{84.31 $\pm$ 0.0000} & 98.85 $\pm$ 0.0003 & 35.65 $\pm$ 0.0086 & 6.6506 $\pm$ 0.0531 & 0.7386 \\
    \cline{2-2}
  Mistral  & MILP & \textbf{68.32 $\pm$ 0.0027} & 
    84.28 $\pm$ 0.0007 & 95.28 $\pm$ 0.0017 & 39.50 $\pm$ 0.0084 & 0.2209 $\pm$ 0.0677 & 0.8516 \\
    \cline{2-2}
  (unseen)  & Z3  &  29.47  $\pm$ 0.0032  &  
    73.07 $\pm$ 0.0081 & 85.90 $\pm$ 0.0035 & 33.79 $\pm$ 0.0164 & 0.0 & 12.8621 \\
    \cline{2-2}
    & \textbf{Ours}  & \underline{ 64.40 $\pm$ 0.0030} & 
    \textbf{84.31 $\pm$ 0.0000} & \textbf{ 98.89$\pm$ 0.0008} & \textbf{ 39.98 $\pm$ 0.0049} & 5.3884 $\pm$ 0.0394 & \textbf{0.6090} \\
    \hline
    \end{tabular}
    }
    \caption {Quantitative comparison of object placement performance. We report the average values and standard deviations for object count (CNT), success rate (SR), scene navigability (NAV), key object navigability (Key\_NAV), out-of-bounds objects (OOB), and placement time per object (PTO) over five runs. Bold text indicates the best performance for each metric.}
    \label{tab:oprl}
\end{table*}
In 3D indoor scene synthesis, LLMs are typically used to generate layout constraints, after which optimization-based solvers are applied to arrange objects and determine their placement within a fixed space. To evaluate the performance of Plan2Place against baseline optimization methods, including Depth First Search (DFS)\cite{Yang_2024_CVPR, pun2025hsm}, Mixed-Integer Linear Programming (MILP) \cite{Yang_2024_CVPR}, and Z3 solver \cite{Wang2026LogicEnvGenTD}, we use 5,200 scenes from the LLaMA-based constraint dataset for training, with the remaining samples reserved for validation. Additionally, Qwen-based and Mistral-based constraint datasets are used as unseen validation sets to further assess generalization. This setup enables us to evaluate the robustness of our model across constraints generated by different LLMs.

 We assess the fidelity of the generated scenes using Object Count (CNT), Success Rate (SR) and plausibility using Scene Navigabilit (NAV), Key Object Navigability (Key NAV), and Object Out-of-Bounds (OOB) as \cite{tam2026sceneeval, Yang_2024_CVPR}, and compare the performance of Plan2Place with the baseline methods. Detailed definitions are provided in Suppl. A-B1

Table \ref{tab:oprl} shows that our method outperforms all baselines on the LLaMA-based constraint dataset, not only in terms of fidelity and most plausibility metrics, but also in average placement time per object (PTO).
For the Mistral-based constraint dataset, although our method performs slightly worse than MILP in terms of CNT and OOB, it still achieves better results on most other metrics and maintains a lower placement time compared to the baselines.
For the Qwen-based constraint dataset, our method remains stable across most metrics, although the Key\_NAV score is slightly lower than that of MILP. In addition, our method requires more time for object placement compared to MILP. This is mainly because a large portion of scenes (approximately 51.6\%), as shown in Suppl. A-A
, contain more than 10 objects, leading to longer action sequences during placement. As a result, our method achieves a significantly higher CNT (27.95\%$\uparrow$) than MILP, which increases the average placement time per scene due to the higher computational cost of sequential decision-making.

Although Z3 achieves the best OOB across all datasets, its CNT is significantly lower than that of most other methods. Overall, across all experiments as shown in Table \ref{tab:oprl}, Plan2Place demonstrates strong stability across metrics and maintains competitive placement time per object on different LLM-based constraint datasets compared to the baselines.

\begin{figure}[ht!]
\begin{subfigure}{0.45\linewidth}
    \includegraphics[scale=0.75]{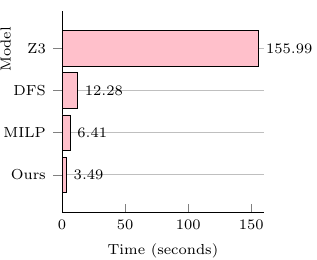}
    \caption{ $\leq$ 10 objects}
    \label{fig:time1}
\end{subfigure}
\begin{subfigure}{0.45\linewidth}
    \includegraphics[scale=0.75]{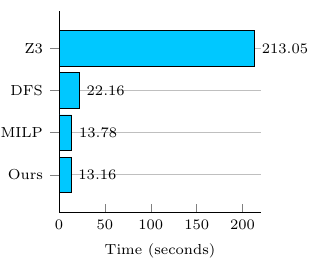}
    \caption{ $>$ 10 objects}
    \label{fig:time1}
\end{subfigure}
 \caption{Comparison of scene completion time between our model and state-of-the-art methods for (a) scenes containing $\leq$ 10 objects and (b) scenes containing $>$ 10 objects.}
  \label{fig:gen_scene_time}
\end{figure}

\textit{Scene Completion Time Comparison.}
We evaluate the average placement time per scene completion between Plan2Place and the baselines on the LLaMA-based constraint dataset. Fig \ref{fig:gen_scene_time} shows that our method significantly outperforms most baselines when the number of objects in a scene is fewer than 10. Although the placement time increases when the number of objects exceeds 10, our method remains faster than the baselines.

\textit{Qualitative Comparison.}
We compare the quality of our generated scenes with other methods based on two criteria: Scene Generation Quality, and Accuracy of VR device localization.  

\textbf{i) Scene Generation Quality}: The quality of the generated scenes is further evaluated through both automated and human evaluations.

\begin{figure}[ht!]
\includegraphics[scale=0.38]{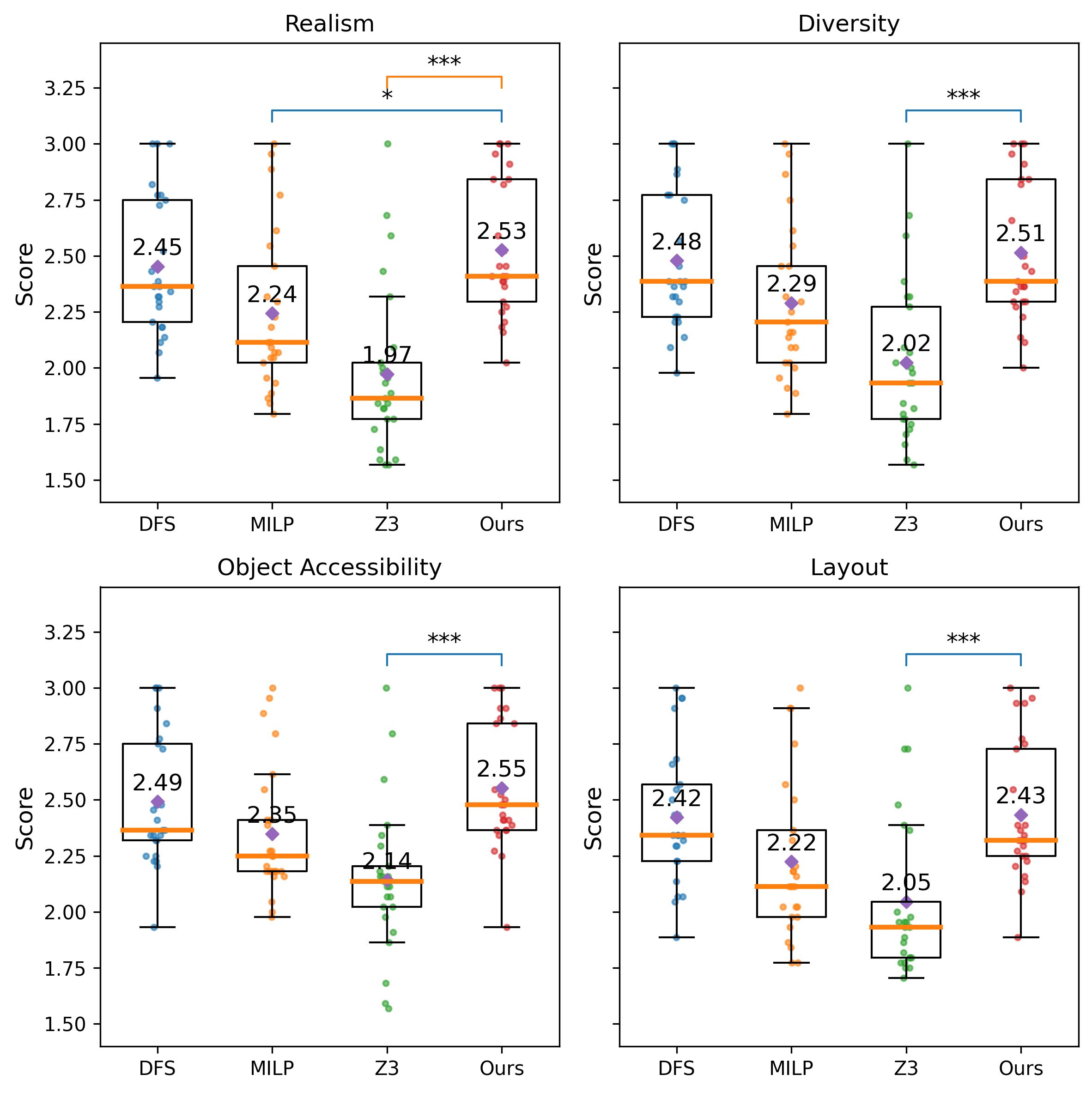}
 \caption{Boxplots of human evaluation results across methods for each metric. Boxes illustrate the interquartile range with bold median lines, and diamonds indicate mean values. Statistical significance is evaluated using Tukey's HSD (*$p < 0.05$, **$p < 0.01$, ***$p < 0.001$). Our method achieves the highest mean scores across all metrics while demonstrating a performance distribution and variability comparable to strong baselines.}
  \label{fig:boxplot}
\end{figure}
\textit{Human Evaluation}: Following \cite{ran2025direct}, we conducted a comprehensive user study with 25 participants on 45 scenes across 9 scene types. For each scene type, we randomly selected two cases from a set of five generated scenes created using the same prompt. In each setting, participants were presented with one set of images generated by different methods. To ensure an unbiased evaluation, the identities of the methods were anonymized.

Participants were asked to rate each generated scene on a three-point scale (1–3), corresponding to poor, moderate, and good, based on the following criteria: realism, diversity, object accessibility, and layout.  To facilitate consistent evaluation, we provided clear instructions for each metric to guide participants in evaluating the different methods, as illustrated in Suppl. A-B2

We conduct one-way ANOVA and Friedman tests to evaluate statistical significance. The results indicate significant differences across methods for all metrics (all $p<0.001$). Specifically, ANOVA presents strong effects (e.g., Realism: $F=13.50$, $p<0.001$), which are further confirmed by the Friedman test (e.g., $\chi^2=57.12$, $p<0.001$), indicating robustness under a non-parametric setting. Moreover, the effect sizes are consistently large across all metrics ($\eta^2$ ranging from 0.19 to 0.30), showing that the choice of method has a substantial impact on human evaluation outcomes.

To further analyze pairwise differences, we perform Tukey's HSD post-hoc tests, as visualized in Fig. \ref{fig:boxplot}. The results show that our method significantly outperforms Z3 across all metrics. Compared to MILP, our method consistently achieves higher scores, with statistically significant improvements observed in Realism, while other differences are not statistically significant. In contrast, no significant difference is observed between our method and DFS across all metrics.

In addition to boxplots, we report the overall score and the distribution of ratings in Table \ref{tab:human_eval_full} to provide a clearer view of reliability. Our method produces fewer low-quality results and more high-quality outcomes, indicating comparable performance with this strong baseline.


\begin{table}[t]
\centering
\resizebox{0.5\textwidth}{!}{
\begin{tabular}{lccccc}
\toprule
Method
& Overall Score 
& Realism (\%) 
& Diversity \%) 
& Accessibility (\%) 
& Layout (\%) \\
\midrule
Z3 
&
2.046
& 38.5 / 25.7 / 35.7 
& 33.8 / 30.0 / 36.2 
& 28.0 / 29.8 / 42.2 
& 32.6 / 30.2 / 37.2 \\
MILP 
&
2.276
& 20.8 / 34.0 / 45.2 
& 16.8 / 37.5 / 45.7 
& 16.2 / 32.9 / 50.9 
& 22.0 / 33.6 / 44.4 \\
DFS
&
\underline{2.461}
& 8.5 / 37.7 / 53.7 
& 5.2 / 41.8 / 53.0 
& 7.5 / 35.9 / 56.6 
& 10.1 / 37.5 / 52.4 \\
\textbf{Ours}  
&
\textbf{2.506}
& 5.1 / 37.3 / 57.6 
& 4.2 / 40.2 / 55.6 
& 5.0 / 34.7 / 60.3 
& 9.1 / 38.5 / 52.4 \\
\bottomrule
\end{tabular}
}
\caption{Human evaluation results with overall score and score distribution (\%) for poor / moderate/ good ratings.}
\label{tab:human_eval_full}
\end{table}
\begin{table}[t]
    \centering
     \resizebox{0.8\columnwidth}{!}{
    \begin{tabular}{|*{5}{c|}}
    \hline
    \textbf{Method} & \textbf{Functional Appropriateness} $\uparrow$ & \textbf{Layout Coherence} $\uparrow$\\ 
    \hline 
    DFS & 3.20 $\pm$ 0.0024 & 
        2.78 $\pm$ 0.0060 \\
    \cline{1-1}
    MILP & 3.05 $\pm$ 0.0517 & 
    2.42 $\pm$ 0.2812 \\
    \cline{1-1}
    Z3  & 2.88 $\pm$ 0.0807 &  
    1.65 $\pm$ 0.3326 \\
    \cline{1-1}
     Ours  & \textbf{3.22 $\pm$ 0.0030} & \textbf{2.81 $\pm$ 0.0089} \\
    \hline
    \end{tabular}
    }
    \caption{Automated evaluation results for Functional Appropriateness and Layout Coherence.}
    \label{tab:autometed_evaluation}
\end{table}
\textit{Automated Evaluation}: To provide a complementary perspective on scene generation quality alongside human evaluation, we employ Qwen2-VL-7B-Instruct as an automated evaluator. It assesses scenes generated by Z3, MILP, DFS, and our method across five independent runs, with 917 images per method. We use two semantic metrics: Functional Appropriateness and Layout Coherence. Detailed definitions are provided in the Suppl. A-B2 
As shown in Table \ref{tab:autometed_evaluation}, our method outperforms the baselines on both metrics.

\textbf{ii) Accuracy of VR device localization:}  
We conducted a user study to evaluate the accuracy of VR device placement across 27 single rooms and 27 apartments. Participants wore a Meta Quest 2 HMD and were asked to verify whether their standing position, from which they observed the virtual environment, corresponded to the anchor object specified in the given instruction. The results achieved an accuracy of 92.59\% for single rooms and 96.30\% for apartments, respectively. Additional visualizations and comparisons with baseline methods are provided in the Suppl. A-F1


\subsubsection{HRI-in-the-loop}
\begin{figure*}[ht!]
\begin{subfigure}{\linewidth}
    \includegraphics[scale=0.157]{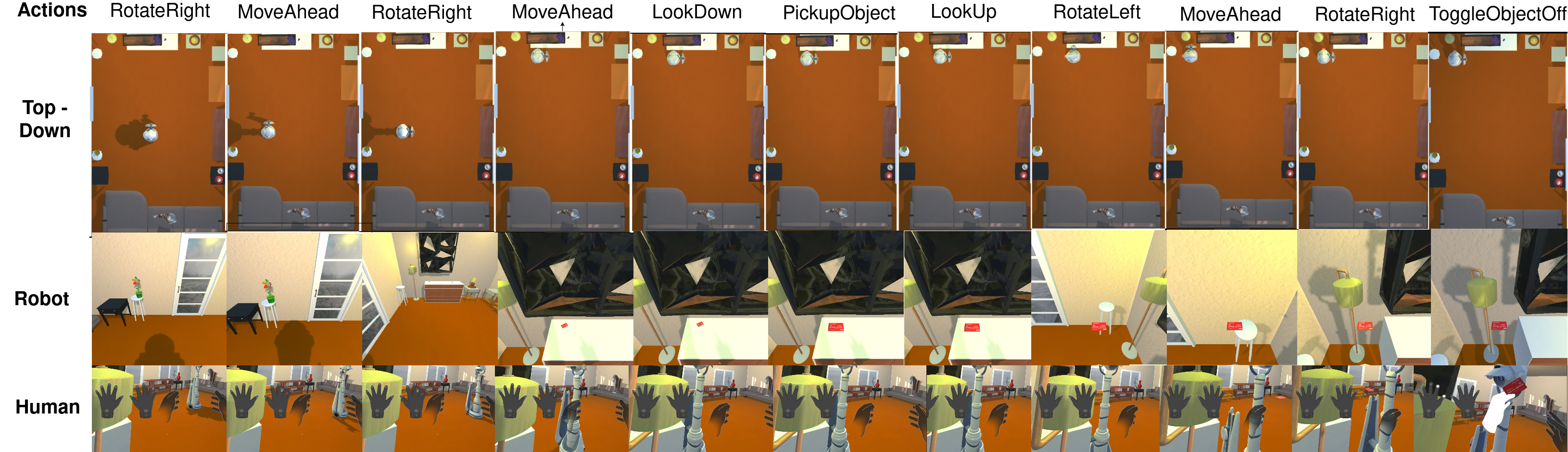}
    \caption{ \textit{Instruction}: Examine a credit card by the light of a floor lamp and then turn it off}
    \label{fig:cool_bread}
\end{subfigure}
\begin{subfigure}{\linewidth}
    \includegraphics[scale=0.094]{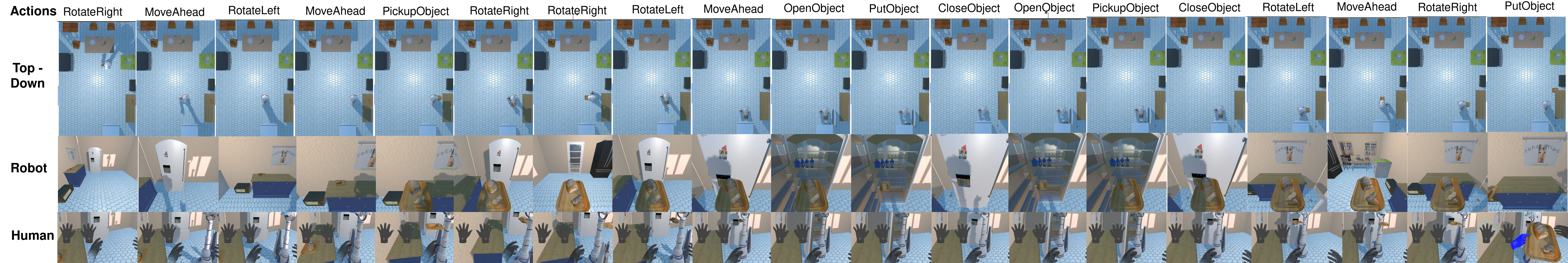}
    \caption{\textit{Instruction}: Put a chilled bread on the counter }
    \label{fig:light_creditcard}
\end{subfigure}
 \caption{Illustration of two representative scenarios of interaction between a robot and a human performing target tasks in a virtual environment. Each scenario is visualized from three distinct perspectives. Top-Down View: Shows the overall layout and the robot's spatial path. Robot's Perspective: Shows the robot's point of view as it navigates and interacts with objects. Human's Perspective: Shows the human's point of view, including interactions with the robot and physical feedback. (a) Lighting Change, and (b) Temperature Change.}
   \label{fig:haptic}
\end{figure*}

Scenario A shown in Fig~\ref{fig:cool_bread} demonstrates a lighting-change task where the system follows the instruction: "Examine a credit card by the light of a floor lamp and then turn it off." The sequence begins as the robot navigates to the table, locates the credit card, and picks it up. It then carries the object to the floor lamp to examine it under the light, and completes the task by turning the lamp off. Throughout this process, the HRI-in-the-loop framework allows the human to interact via a haptic glove and perceive environmental changes; they feel object vibrations when touching the credit card, while the glove's color shifts from black to white to reflect the change in lighting conditions. Scenario B shown in Fig~\ref{fig:light_creditcard} demonstrates a temperature-change task where the system follows the instruction: "Put a chilled bread on the counter." The sequence begins as the robot navigates to the bread, picks it up, and then moves to the refrigerator to cool or chill the item. Finally, the robot places the chilled bread onto the counter. Throughout the process, the HRI-in-the-loop framework enables the user to interact through a head-mounted display and a haptic glove, allowing them to grasp the bread and feel its temperature change via vibration feedback. As the temperature drops, the color of the glove shifts from black to blue to reflect the cooling effect.

The scenarios illustrate that the model can successfully generate immersive 3D indoor environments integrated with a haptic device, enabling effective human-robot interaction. Users can actively perceive environmental changes such as variations in light, temperature, and object interactions through multimodal feedback, significantly enhancing overall realism and the user experience.

\subsection{Ablation Study}
Through a series of ablation studies, we systematically evaluated the effectiveness of each component in our framework.
\subsubsection{Evaluation of Feature Integration Strategies in MSCF} To assess the effectiveness of various feature integration strategies within our MSCF module, we conducted a comparative study of three mechanisms: concatenation, self-attention, and cross-attention. As shown in Table \ref{tab:mechanism}, the cross-attention mechanism outperforms the other methods overall.
\begin{table}[ht!]
    \centering
    \centering
     \resizebox{1\columnwidth}{!}{
    \begin{tabular}{|*{7}{c|}}
    \hline
    \textbf{Mechanism} & \textbf{CNT} $\uparrow$ & \textbf{NAV} $\uparrow$ & \textbf{Key\_NAV} $\uparrow$ & \textbf{OOB} $\downarrow$  \\ 
    \hline 
    Concatenation & \textbf{80.74 $\pm$ 0.0012} & 99.91 $\pm$ 0.0003 & 55.17 $\pm$ 0.0043 & 6.6306 $\pm$ 0.0115\\
    
    Cross-Attention  & \underline{80.68 $\pm$ 0.0011} & \textbf{99.92 $\pm$ 0.0003} & \textbf{56.79 $\pm$ 0.0053} & \underline{6.6232 $\pm$ 0.0216} \\
    
    Self-Attention & 80.63 $\pm$ 0.0009 & \textbf{99.92 $\pm$ 0.0003} & \underline{55.57 $\pm$ 0.0083} & \textbf{6.6037 $\pm$ 0.0108}\\

    \hline
    \end{tabular}
    }
     \caption{Impact of different feature integration methods in MSCF.}
    \label{tab:mechanism}
\end{table}

\subsubsection{Effect of SFF}
\begin{table}[ht!]
    \centering
    \centering
     \resizebox{1\columnwidth}{!}{
     \begin{tabular}{|*{7}{c|}}
    \hline
    \textbf{Mechanism}& \textbf{Lay} $\uparrow$ & \textbf{NAV} $\uparrow$ & \textbf{Key\_NAV} $\uparrow$ & \textbf{OOB} $\downarrow$  \\ 
    \hline 
    w/o SFF & 3.18 & 99.90 $\pm$ 0.0004 & 55.05 $\pm$ 0.0062 & \textbf{6.6279 $\pm$ 0.0028}\\
    SFF & \textbf{3.22} & \textbf{99.91 $\pm$ 0.0003} & \textbf{55.17 $\pm$ 0.0043} & 6.6306 $\pm$ 0.0115\\
    \hline
    \end{tabular}
    }
     \caption{Impact of of SFF module on Plan2Place performance.}
    \label{tab:sffm}
\end{table}
To further analyze the effectiveness of the SFF in our method, we conducted an ablation study comparing two configurations: with and without SFF. We first evaluated layout coherence on approximately 100 generated images for both configurations. As shown in Table \ref{tab:sffm}, incorporating the SFF module improves overall performance by 4\% compared to the variant without SFF. Although the out-of-bounds (OOB) rate increases slightly, our method still achieves superior performance on navigability-related metrics in large-scale experiments. This behavior can be attributed to the fact that SFF encourages tighter, more semantically meaningful object groupings, which may occasionally position objects closer to scene boundaries. Importantly, the overall improvement in navigability metrics indicates that SFF achieves a better balance between spatial efficiency and functional accessibility, suggesting that the learned representations are not only more coherent but also more practical for real-world scene synthesis.

\subsubsection{Effect of Auxiliary loss}

We conducted an ablation study to analyze the effectiveness of the auxiliary loss in Plan2Place. Specifically, we compared two variants using the identical training configuration: one trained with the auxiliary loss in the policy learning process, and one without.

As shown in Table \ref{tab:auxiliary_loss}, incorporating the auxiliary loss improves layout coherence and fidelity metrics. For plausibility metrics, both the NAV and OOB scores are slightly lower compared to the model without the auxiliary loss. This can be attributed to the model placing more objects when the auxiliary loss is applied, which results in denser layouts. Nevertheless, navigability with respect to key objects (Key\_NAV) improves, which aligns directly with the primary objective of our problem.

\begin{table}[ht!]
    \centering
    \centering
     \resizebox{\columnwidth}{!}{
    \begin{tabular}{|*{9}{c|}}
    \hline
     \textbf{Loss} & \textbf{Lay} $\uparrow$ & \textbf{CNT} $\uparrow$ & \textbf{NAV} $\uparrow$ & \textbf{Key\_NAV} $\uparrow$ & \textbf{OOB} $\downarrow$  \\ 
    \hline 
    w/o Aux &  
    3.2766 & 80.65 $\pm$ 0.0010 & \textbf{99.94 $\pm$ 0.0001} & 56.22 $\pm$ 0.0083 & \textbf{6.6161 $\pm$ 0.0227}\\
    + Aux &  
    \textbf{3.3050} & \textbf{80.68 $\pm$ 0.0011} & 99.92 $\pm$ 0.0003 & \textbf{56.79 $\pm$ 0.0053} & 6.6232 $\pm$ 0.0216 \\
    \hline
    \end{tabular}
    }
     \caption{Influence of Auxiliary Loss on Plan2Place Performance}
    \label{tab:auxiliary_loss}
\end{table}

\begin{table}[ht]
\centering
{\normalsize
\resizebox{0.5\textwidth}{!}{
    \begin{tabular}{|c|c|c|c|c|c|}
    \hline
         & Implicit  & Closed-Loop  &  Adaptive  & Immersive  & Real-time \\
         Method & Object & User & Layout&  VR/Haptic & Interaction\\
         & Reasoning & Interaction & Optimization&  Support & Adaptation\\
        \hline 
        Holodeck\cite{Yang_2024_CVPR} &$\bigtriangleup$ & \xmark & $\bigtriangleup$ & \xmark & \xmark\\
         LayoutGPT\cite{feng2024layoutgpt} &\xmark & \xmark & $\bigtriangleup$ & \xmark & \xmark \\
          
         Ours & \tick & \tick & \tick & \tick & \tick\\
         \hline
    \end{tabular}
    }}
     \caption{Conceptual comparison with representative prior methods across generation, interaction, and optimization capabilities. $\bigtriangleup$ denotes partial support, \protect\ding{51}~ indicates the method supports the property, and \protect\ding{55}~ indicates it does not.}
\label{tab:comparision_environment}
\end{table}

\subsection{Discussion}
The results of this study highlight the importance of integrating content generation and user interaction within a unified framework. While recent advances in language-driven scene generation have demonstrated strong capabilities in semantic understanding and layout synthesis, these approaches typically operate in a feedforward manner without considering how generated content is perceived and interacted with by users. Our findings suggest that this separation limits both the usability and realism of generated environments. By explicitly modeling a closed-loop interaction between generation, perception, and action, the proposed framework enables generated environments to adapt dynamically to user behavior. This interaction-driven refinement leads to improved immersion, interaction quality, and task efficiency, as demonstrated in both benchmark evaluations and user studies. In particular, the integration of multimodal signals—including visual, spatial, and haptic feedback—plays a key role in aligning generated content with human perception, highlighting the importance of multimodal reasoning in interactive multimedia systems. The comparison with prior methods further illustrates the fragmented nature of existing approaches. As shown in Table \ref{tab:comparision_environment}, previous work has largely focused on individual aspects such as scene generation or layout optimization, while lacking mechanisms for closed-loop interaction and immersive feedback. In contrast, the proposed framework unifies these components within a single pipeline, enabling continuous adaptation of generated environments based on user interaction. \\
\hspace*{10px}Despite these advantages, several limitations remain. First, the current framework is evaluated primarily in indoor environments with relatively structured layouts, and its scalability to more complex or open-world scenarios requires further investigation. Second, while the system incorporates visual and haptic feedback, additional modalities such as audio or social interaction cues could further enhance realism and immersion. Finally, the computational cost associated with reinforcement learning-based optimization may limit real-time deployment in large-scale environments. Overall, these findings suggest that future multimedia systems should move beyond static content generation toward adaptive frameworks in which generation, perception, and interaction are jointly optimized. Closing this loop is essential for enabling more responsive, immersive, and human-centered multimedia experiences.

\section{Conclusion}
In this paper, we presented a unified framework that closes the loop between language-driven 3D scene generation and immersive user interaction. By integrating large language models, reinforcement learning, and an HRI-in-the-loop paradigm within virtual reality, the proposed approach enables the automatic generation of interactive 3D environments that continuously adapt to user perception and behavior. Unlike conventional methods that treat content generation and interaction as separate processes, our framework explicitly couples generation, perception, and action within a single multimodal pipeline. This design not only improves scene generation quality but also enhances usability, realism, and responsiveness in interactive environments. Extensive experiments on the ALFRED benchmark demonstrated state-of-the-art performance in task-based scene generation, while user studies confirmed consistent improvements in immersion, interaction quality, and task efficiency. These results highlight the practical value of integrating multimodal interaction into generative systems. Overall, this work advances multimedia systems by integrating content generation, user perception, and human-robot interaction into a unified closed-loop framework. We believe that such tightly coupled systems represent a promising direction for next-generation multimedia applications, enabling more adaptive, immersive, and human-centered experiences.

\section*{Acknowledgment}
This work was supported by the Information Technology Research Center (ITRC) support program (IITP-2026-RS-2022-00156354) and a Korean government grant (MSIT) (No.RS-2019-II190231) from the Institute of Information \& Communications Technology Planning \& Evaluation (IITP) as well as by the Basic Science Research Program through the National Research Foundation of Korea (NRF) funded by the Ministry of Education (2020R1A6A1A03038540).

\ifCLASSOPTIONcaptionsoff
  \newpage
\fi



%


\bibliographystyle{IEEEtran}
\bibliography{bibtex/bib/references}

%

\enlargethispage{-1.4in}
\newpage
 \appendices
 \section{Supplementary}

\subsubsection{Reward Energy Function}
\label{label:reward_details}
To provide informative feedback to the agent, we  formulate the reward function as an energy minimization objective instead of relying on sparse or rule-based signals. The energy function measures the degree of constraint violation in the generated layout, including relational consistency, interaction affordances, navigation feasibility, object overlap, and out-of-bound placement. The reward is defined as the negative energy, such that maximizing the reward is equivalent to
minimizing the overall energy, thereby encouraging physically
valid and semantically coherent layouts.

\textit{Relational Energy}: A metric that quantifies the consistency of object spatial relationships against LLM-derived, $E_{i,j}^{geom}$. To mitigate potential bias given by these constraints, we incorporate prior knowledge extracted from
human-designed datasets, such as 3D-FRONT and ProcTHOR, $E_{i,j}^{prior}$. This prior information enables the agent to capture
aesthetic and common-sense placement patterns that align with
human-designed environments. 
\begin{equation}
\begin{aligned}
E_{rel} = log(1 + \frac{1}{N_c} \sum E_{ij}^{geom}(c).(1 + \lambda \hat{E}_{i,j}^{prior}(c))_{(i,j,c) \in C}) \\
\text{Geometric Energy: } E_{i,j}^{geom}(c) = E^c(i,j) \\
\text{Data Prior: } \hat{E}_{i,j}^{prior} = clip(\frac{E_{i,j}^{prior}}{E_{max}}) 
\end{aligned}
\end{equation}
Here, $C$ is the set of LLM-based constraints.

\textit{Collision Energy}: A metric that quantifies the extent of overlap between the current object and previously placed objects in the environment.
\begin{equation}
\begin{aligned}
E_{collision} = log(1 + \sum_{i <j} (area(P_i \bigcap P_j))^2)
\end{aligned}
\end{equation}
where $P_i$ and $P_j$ represent the polygons corresponding to the placed object $i$ and the current object $j$, respectively, derived from their bounding boxes.

\textit{Out of Bound Energy} ($E_{oob}$): A metric that quantifies the proportion of an object’s area that lies outside the room boundary. 
\begin{equation}
\begin{aligned}
\rho_{oob}^i = \frac{area (P_i \bigcap p_r)}{area(P_i) + \epsilon} = log(1 + \sum_{i\in \mathbf{O}}(1 - \rho_{oob}^i)^2)
\end{aligned}
\end{equation}
where $P_r$ denotes the room polygon, and let $\rho_{oob}^i$ represent the overlap ratio between the room and the object $i$ being considered, and $\epsilon$ is set to 1e-8.

\textit{Affordance Energy}: A metric that quantifies functional usability through the affordance-aware reward function. While prior works primarily focus on spatial feasibility, they often overlook whether the generated layouts support realistic human-object interactions. To address this limitation, we design an affordance-based energy term that encourages the preservation of accessible interaction space for functional objects.

A clearance region $\mathcal{C}_i$ is defined in front of the object to
represent the required free space for interaction (e.g., opening
a refrigerator door). It is constructed as a rectangular area
aligned with the object’s orientation, with predefined depth
and lateral margin.

\begin{equation}
\begin{aligned}
E_{aff} = \lambda_1 \mathbb{I}(\mathcal{C}_i \nsubseteq P_r) + \lambda_2.\sum_{j\neq i}\mathbb{I}(\mathcal{C_i \cap  P_j \neq \emptyset})
\end{aligned}
\end{equation}
where $\mathcal{P}_r$ denotes the room polygon, $\mathcal{P}_j$ represents the polygon of other objects, and $\mathbb{I}$ is the indicator function.

This formulation penalizes placements that violate accessi-
bility constraints, either by placing the clearance region outside
the room or by obstructing it with other objects. As a result,
the agent is encouraged to generate layouts that not only
satisfy geometric validity but also support realistic interaction
affordances.

\textit{Navigation Energy}: A metric that quantifies the accessibility of placed objects from the robot’s position within the environment. In this study, an A* search algorithm is employed to compute the shortest path from the robot’s initial position, typically located near the main entrance, to each target object. Based on the feasibility of these paths, we determine the set
of reachable targets.

\begin{equation}
\begin{aligned}
\text{Reachability Ratio}: \rho_{nav} = \frac{\text{reachable targets}}{\text{total targets}} \\
E_{nav} = (1 - \rho_{nav})^2
\end{aligned}
\end{equation}

\subsection{Dataset}
\label{label:dataset}
\begin{figure}[ht!]
\centering
\begin{subfigure}{0.39\linewidth}
    \includegraphics[scale=0.63]{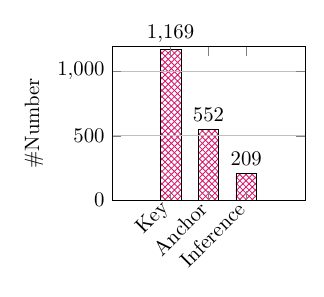}
    \caption{}
    \label{fig:data1}
\end{subfigure}
\begin{subfigure}{0.27\linewidth}
\centering
    \includegraphics[scale=0.29]{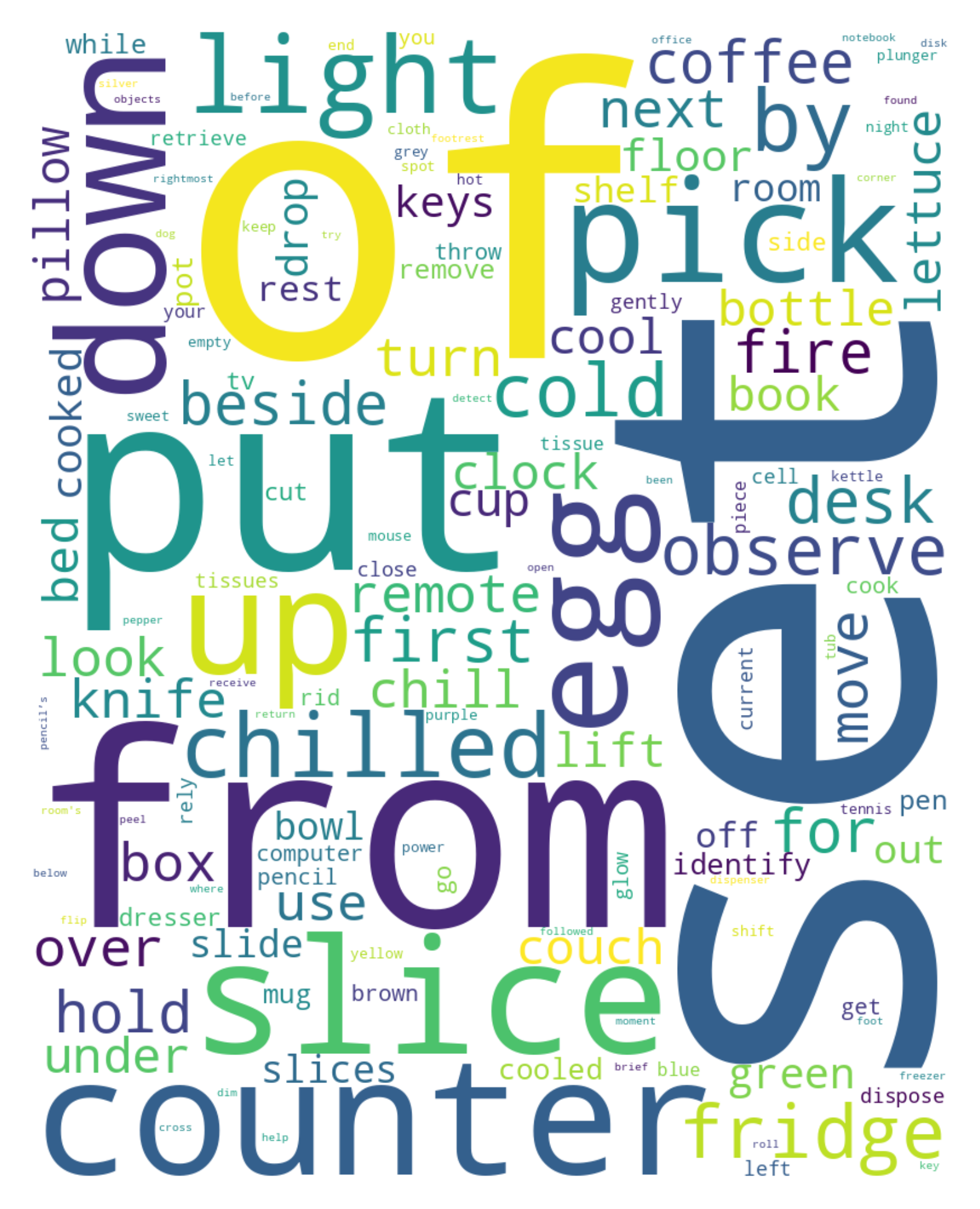}
    \caption{ }
    \label{fig:template_unseen}
\end{subfigure}
\begin{subfigure}{0.27\linewidth}
    \includegraphics[scale=0.29]{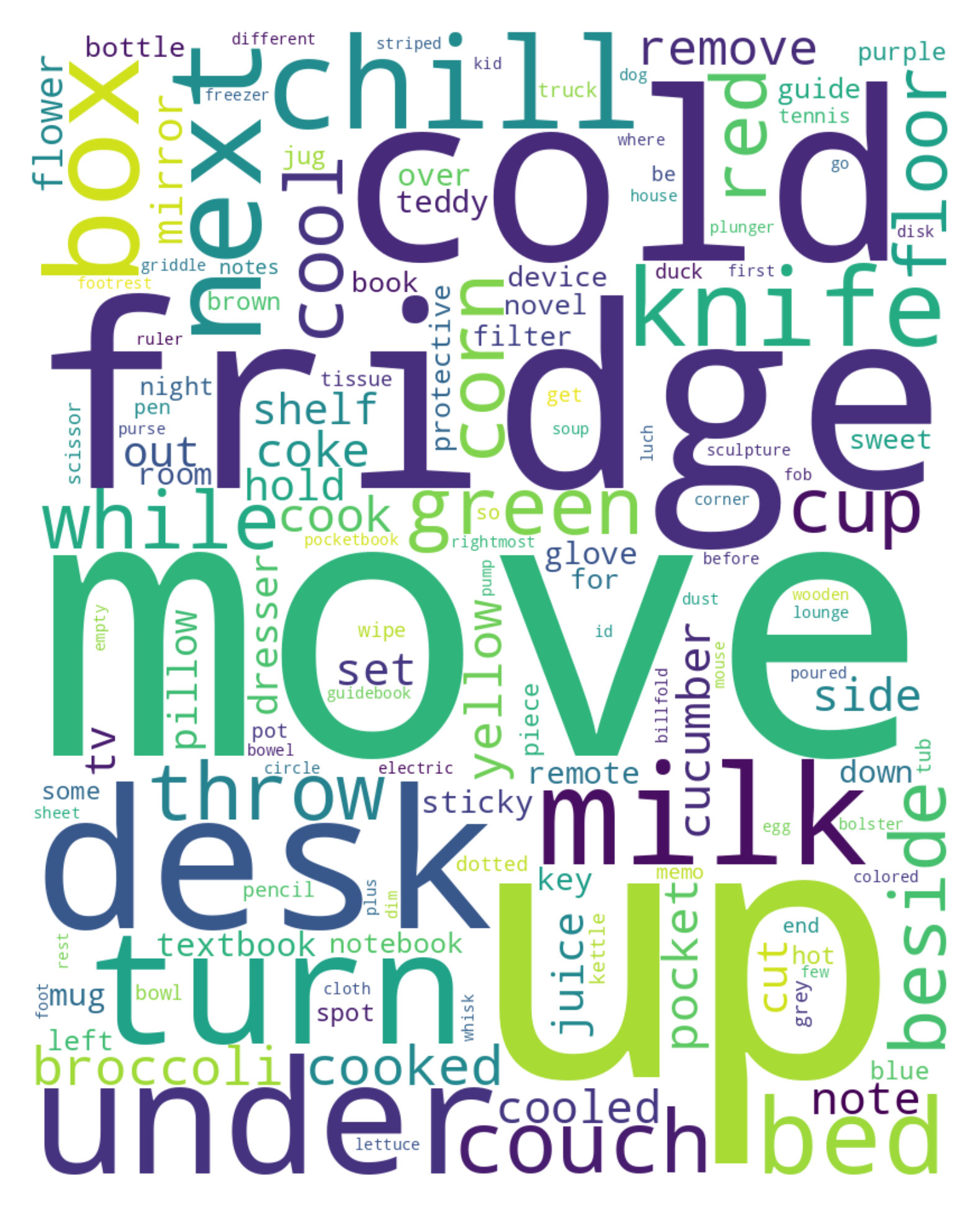}
    \caption{ }
    \label{fig:ood_unseen}
\end{subfigure}
\caption{Overview of the task-based dataset: (a) distribution of object categories in the ground-truth scene graphs; (b–c) word cloud visualizations of the TS and OS datasets.}
\label{fig:task_data}
\end{figure}

Figure \ref{fig:task_data}(b–c) provides detailed statistics of the TS and OS sets, respectively. The distribution of object categories, including key, anchor, and inference objects, is shown in Figure \ref{fig:data1}. 

\begin{figure}[ht!]
\begin{subfigure}{0.3\linewidth}
    \includegraphics[scale=0.4]{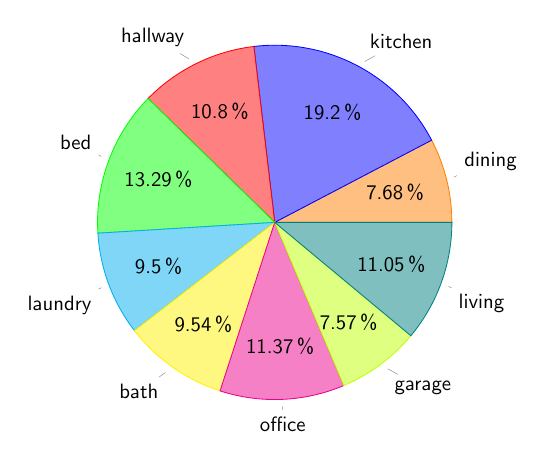}
    \caption{}
    \label{fig:llama-db}
\end{subfigure}
\hfill
\begin{subfigure}{0.3\linewidth}
    \includegraphics[scale=0.4]{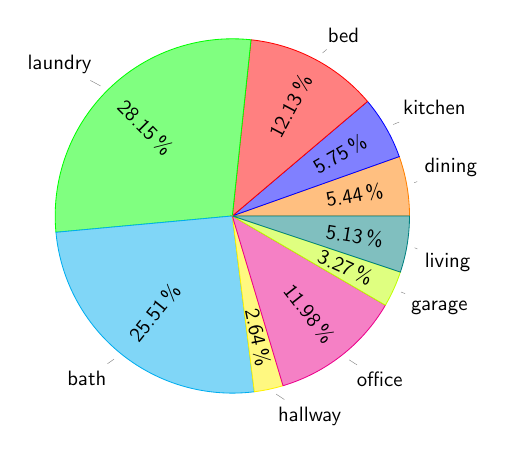}
    \caption{ }
    \label{fig:mistral-db}
\end{subfigure}
\hfill
\begin{subfigure}[b]{0.3\linewidth}
    \includegraphics[scale=0.4]{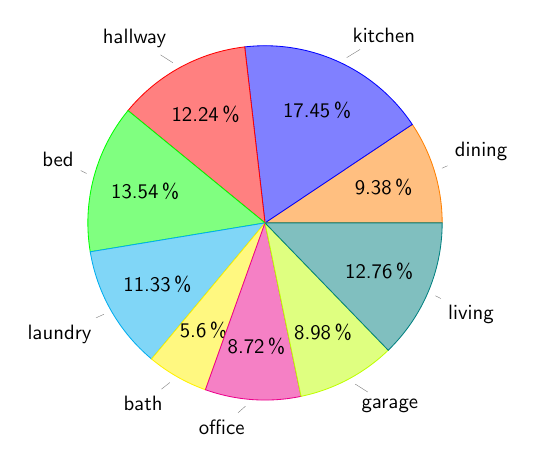}
    \caption{ }
    \label{fig:qwen-db}
\end{subfigure}
\begin{subfigure}{0.33\linewidth}
\centering
    \includegraphics[scale=0.4]{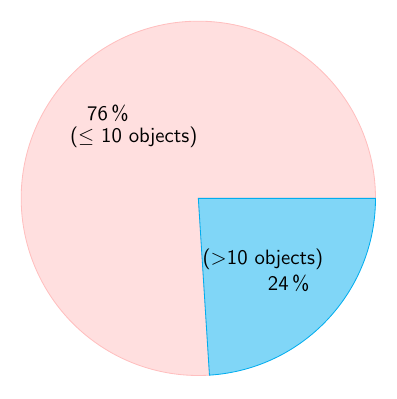}
    \caption{ }
    \label{fig:llama-db_num}
\end{subfigure}
\hfill
\begin{subfigure}{0.3\linewidth}
\centering
    \includegraphics[scale=0.4]{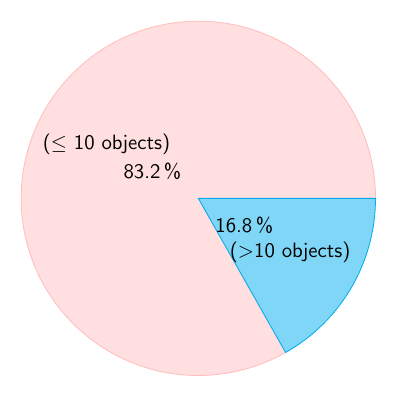}
    \caption{ }
    \label{fig:mistral-db_num}
\end{subfigure}
\hfill
\begin{subfigure}{0.3\linewidth}
\centering
    \includegraphics[scale=0.4]{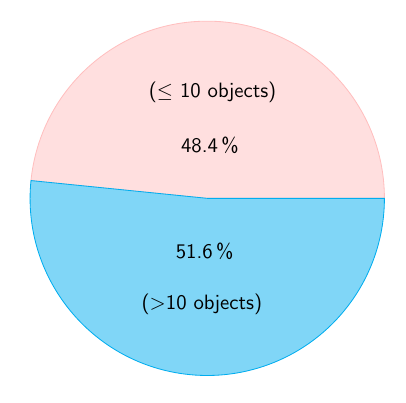}
    \caption{ }
    \label{fig:qwen-db_num}
\end{subfigure}
\hfill
 \caption{Illustration of the LLM-based constraint datasets: (a) LLaMA, (b) Mistral, and (c) Qwen. The distribution of object counts is further divided into two groups ($\leq$ 10 and $>$ 10), as shown in (d)-(f).}
  \label{fig:llmdataset}
\end{figure}

Building on prior work \cite{Yang_2024_CVPR}, we adopt their prompts and apply them to inputs from the ALFRED dataset, pairing them with nine representative room categories across various LLMs, including LLaMA, Mistral, and Qwen. After filtering, we retain only the scenes that contain objects corresponding to the input instructions.  Fig \ref{fig:llmdataset} illustrates the distribution of scene categories and the percentage of scenes with more than 10 objects and those with 10 or fewer objects across different LLMs. 

\subsection{Evaluation Metrics}
\textbf{- Language-Driven Scene Representation}
\label{label:languagedriven_SR_metric}
We utilize Graph Edit Distance (GED) \cite{wang2021combinatorial, chen2019efficient}, Instruction Recall (iRecall) \cite{lin2024instructscene}, and F1 score \cite{jiao2023instruct}. Specifically, GED quantifies the similarity (or dissimilarity) between two graphs. iRecall measures the proportion of required triplets—(subject, relation, object)—that are successfully generated in the scene relative to those specified in the instructions. This metric simultaneously considers all three components of a layout relation. Finally, the F1 score evaluates whether the objects generated by the LLM are consistent with the ground-truth object. 

\textbf{- Position Prediction of VR Devices}
\label{label:vr_metric}
To provide a comprehensive evaluation of object category prediction in the Position Prediction of VR Devices, we assess the model using standard metrics, including accuracy, precision, recall, and F1-score.

\textbf{- Object Placement Optimization} 

\subsubsection{Details of the quantitative metrics} We utilize fidelity and plausibility metrics for the quantitative evaluation.
\label{label:opo_metric}

i) \textit{Fidelity Metrics}: 
\begin{itemize}
\item \textbf{Object Count (CNT)} check if the number of objects in the generated scene matches the quantities specified in the list of input objects.
\item \textbf{Success Rate (SR)} whether the quantities of anchor and inference objects match those specified in the task description. 
\end{itemize}

ii) \textit{Plausibility Metrics}: 
\begin{itemize}
\item \textbf{Scene Navigability (NAV)} measures whether the object arrangement provides continuous space for movement. Free space is defined as the floor area not occupied by objects or interior elements. Navigability is evaluated as the ratio of the largest connected free space to the total free space. This is computed by projecting the scene onto a 2D occupancy mask and applying connected component analysis. While this metric captures movement at the room level, it fails to ensure that individual objects remain accessible or usable.

\item \textbf{Key Object Navigability (Key\_NAV)} addresses this limitation by measuring whether the arrangement of anchor and inferred objects in the task description provides sufficient connected free space for navigation. We construct an occupancy grid to represent the free space in the room (i.e., areas not occupied by objects), and then apply A* search to identify paths from the agent’s initial position (e.g., the main door) to the anchor and inferred objects. This allows us to evaluate whether the objects specified in the input instruction are reachable and usable by the agent. 

\item \textbf{Object Out-of-Bounds (OOB)} ensures that objects remain within the boundaries of the scene’s floor plan.
\end{itemize}
\subsubsection{Details of Semantic Quality}
\label{label:opo_qualitive_metric}

We provided clear instructions for each metric to guide participants in evaluating the different methods, as follows: 
\begin{itemize}
 \item \textbf{Realism} evaluates the realistic and plausibility of the generated scene. 
 \item \textbf{Diversity} evaluates the different and varied generated scenes across different generated scenes of the same method.  
 \item \textbf{Object Accessibility} ensures that the functional sides of objects are accessible. 
 \item \textbf{Layout} evaluates the quality of spatial organization in terms of structure, spacing, flow, and object grouping.
\end{itemize}
The detailed metrics are used in automated evaluation:
\begin{itemize}
\item \textbf{Functional Appropriateness}, which evaluates whether object types are suitable for the scene context. 
\item \textbf{Layout Coherence}, which measures the consistency of spatial and functional relationships among objects.
\end{itemize}
\subsection{Experiment Details}
\label{label:exp_details}
\subsubsection{Experiment Setting for Language-Driven Scene Representation}
During training, we use a learning rate of 2e-5 for 50 epochs, with the Adam optimizer.





\subsubsection{Experiment Setting for Position Prediction}
We use a batch size of 16, train for 50 epochs, and set the learning rate to $3e^{-4}$. The Adam optimizer is used for training. 
\subsubsection{Experiment Setting for Plan2Place}
The model is trained for 50 epochs with a batch size of 1 and a learning rate of $1 \times 10^{-4}$. The discount factor $\gamma$ is set to 0.99, and optimization is performed using the Adam optimizer.

\textbf{Baselines}
\label{label:plan2place}

- \textit{DFS and MILP solver}. These methods are employed similarly to those in \cite{Yang_2024_CVPR}.

- \textit{Z3 solver}\footnote{https://github.com/Z3Prover/z3}. The 3D positions and orientations of objects, along with the doorway, are treated as variables, with constraints defined consistently across DFS, MILP, and our method.

\subsection{More Results for Position Prediction of VR Devices}
\label{label:pos_pre_vr_supp}
\subsubsection{Evaluation across Standard Metrics}
\begin{figure}[ht!]
\begin{subfigure}{0.32\linewidth}
    \includegraphics[scale=0.55]{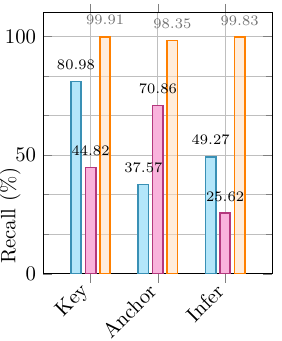}
    \caption{}
    \label{fig:re-t1}
\end{subfigure}
\hfill
\begin{subfigure}{0.32\linewidth}
    \includegraphics[scale=0.55]{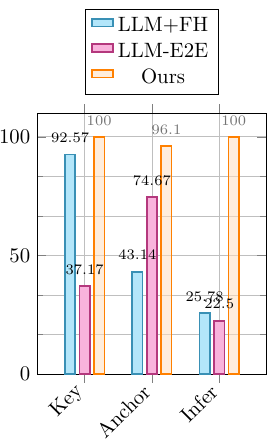}
    \caption{ }
    \label{fig:re-t2}
\end{subfigure}
\centering
\begin{subfigure}{0.32\linewidth}
    \includegraphics[scale=0.55]{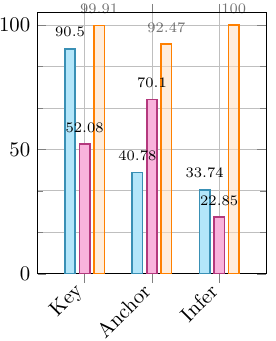}
    \caption{ }
    \label{fig:re-t3}
\end{subfigure}

\begin{subfigure}{0.32\linewidth}
    \includegraphics[scale=0.55]{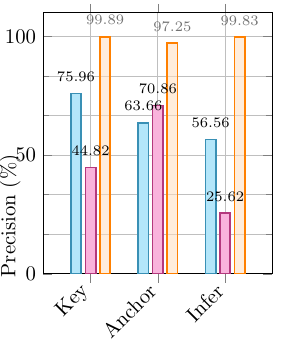}
    \caption{}
    \label{fig:pre-t1}
\end{subfigure}
\hfill
\begin{subfigure}{0.32\linewidth}
    \includegraphics[scale=0.55]{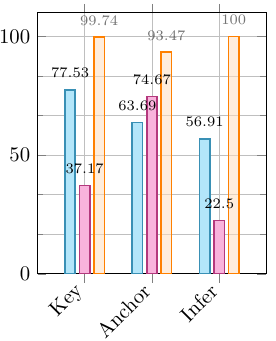}
    \caption{ }
    \label{fig:pre-t2}
\end{subfigure}
\centering
\begin{subfigure}{0.32\linewidth}
    \includegraphics[scale=0.55]{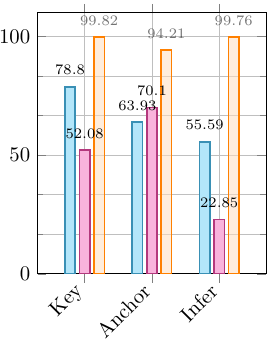}
    \caption{ }
    \label{fig:pre-t3}
\end{subfigure}

\begin{subfigure}{0.32\linewidth}
    \includegraphics[scale=0.55]{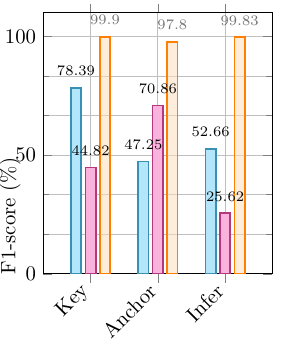}
    \caption{}
    \label{fig:f1-t1}
\end{subfigure}
\hfill
\begin{subfigure}{0.32\linewidth}
    \includegraphics[scale=0.55]{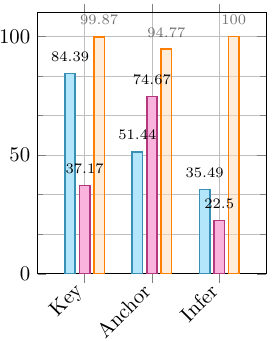}
    \caption{ }
    \label{fig:f2-t2}
\end{subfigure}
\centering
\begin{subfigure}{0.32\linewidth}
    \includegraphics[scale=0.55]{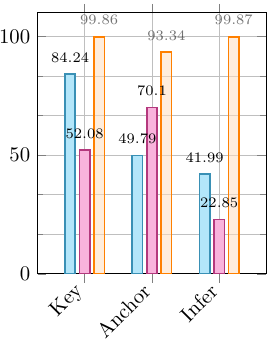}
    \caption{ }
    \label{fig:f1-t3}
\end{subfigure}
 \caption{Comparison of VR Device Position Prediction Performance, measured by recall, precision, and F1-score, across different datasets: (a, d, g) ID, (b, e, h) TS, and (c, f, i) OS}
  \label{fig:vr_position}
\end{figure}

Fig. \ref{fig:vr_position} further shows a comparison between our model, LLM-E2E, and LLM+FH. The results indicate that our model consistently outperforms both baselines across all evaluation sets in terms of F1-score, precision, and recall.

\subsubsection{Effect of Hidden Dimensionality on Position Prediction of VR Devices Performance}
\label{label:vr-aware_ablation_study}
Table \ref{tab:llm-pos-haptic_abla} presents a comparative analysis of VR position prediction performance across different hidden dimensions for both token embedding and MLP layer. Experiments are conducted on three evaluation sets, including ID, TS, and OS. The results consistently demonstrate that a hidden dimension of 128 achieves stability and effectiveness compared to larger configurations (256 and 512) across most datasets.   
\begin{table}[ht!]
    \centering
     \resizebox{\columnwidth}{!}{
    \begin{tabular}{|*{8}{c|}}  
    \hline
    \multirow{2}{*}{\textbf{Dataset}} & \multirow{3}{*}{\textbf{Class}} & \multicolumn{3}{|c|}{\textbf{Acc} $\uparrow$}\\ 
    \cmidrule{3-5}
     & & 128-D & 256-D & 512-D\\ 
    \hline
   ID & Key & \textbf{99.94 $\pm$ 0.0000} &  99.87 $\pm$ 0.0002 & \textbf{99.94 $\pm$ 0.0000}\\
   \cline { 2 - 2}
   & Anchor & \textbf{98.79 $\pm$ 0.0000} & 98.24 $\pm$ 0.0085 & \textbf{ 98.79 $\pm$ 0.0000}\\
    \cline { 2 - 2}
   & Inference & \textbf{100.0 $\pm$ 0.0000} & 99.89 $\pm$ 0.0003 &  \textbf{ 100.0$\pm$ 0.0000}\\
   \hline
   TS & Key & \textbf{99.87 $\pm$ 0.0000} & 99.81 $\pm$ 0.0000 & 99.80 $\pm$ 0.0000\\
   \cline { 2 - 2}
   & Anchor & \underline{92.65 $\pm$ 0.0006} & \textbf{95.65 $\pm$ 0.0003} & 92.43 $\pm$ 0.0004\\
   \cline { 2 - 2}
   & Inference & 99.93 $\pm$ 0.0000 & \textbf{100.0 $\pm$ 0.0000} & 99.93$\pm$ 0.0000\\
   \hline
   OS & Key & \textbf{99.94 $\pm$ 0.0000} & 99.80 $\pm$ 0.0001 &  99.80 $\pm$ 0.0003\\
     \cline { 2 - 2}
   & Anchor & \underline{94.57 $\pm$ 0.0005} & \underline{94.57 $\pm$ 0.0003} & \textbf{ 94.68 $\pm$ 0.0005}\\
    \cline { 2 - 2}
   & Inference & 99.94 $\pm$ 0.0000 & 99.94 $\pm$ 0.0000 & 99.94 $\pm$ 0.0000\\
    \hline
    \end{tabular}
    }
    \caption{Comparison of object category prediction performance under varying hidden dimensions}
    \label{tab:llm-pos-haptic_abla}
\end{table}

\subsection{More Results for Object Placement Optimization}

\subsubsection{Effect of Hidden dimension for Plan2Place}
We analyze the impact of hidden dimensionality on Plan2Place's performance under multiple evaluation metrics. As shown in Table \ref{tab:vary_hidden_dim}, no single configuration consistently outperforms others across all metrics, revealing a trade-off between layout quality, navigability, and object placement capacity.

Specifically, a hidden dimension of 128 achieves the best performance in terms of layout coherence and key object navigability, indicating its effectiveness in capturing relational structure and task-relevant interactions. In contrast, increasing the hidden dimension to 256 allows the model to place more objects and reduces out-of-bound errors (OOB), suggesting improved capacity for handling denser scenes. Meanwhile, a smaller hidden dimension of 64 yields competitive performance across most metrics, particularly in overall object navigability, while maintaining second-best results in layout coherence, key object navigability, and OOB.

Considering the primary objective of our task, ensuring coherent layouts and reliable accessibility of key objects. Here, we select a hidden dimension of 128 as the default setting, as it provides the best balance between structural consistency and task-oriented performance.    
In addition, these observations suggest that increasing model capacity does not necessarily lead to better relational reasoning, but instead primarily improves spatial feasibility in dense scenarios.
\begin{table}[ht!]
    \centering
    \centering
     \resizebox{\columnwidth}{!}{
    \begin{tabular}{|*{8}{c|}}
    \hline
    \textbf{Hidden dim} & \textbf{Lay} $\uparrow$ & \textbf{CNT} $\uparrow$ & \textbf{NAV} $\uparrow$ & \textbf{Key\_NAV} $\uparrow$ & \textbf{OOB} $\downarrow$ \\ 
    \hline
    64 & 3.21 & 80.75 $\pm$ 0.0013  & \textbf{99.94 $\pm$ 0.0001} & 56.16 $\pm$ 0.0058 & 6.6122 $\pm$ 0.0182\\
    128 & \textbf{3.30} & 80.68 $\pm$ 0.0011 & 99.92 $\pm$ 0.0003 & \textbf{56.79 $\pm$ 0.0053} & 6.6232 $\pm$ 0.0216 \\
    256 & 3.18 & \textbf{80.81 $\pm$ 0.0007} & 99.92 $\pm$ 0.0004 & 55.94 $\pm$ 0.0072 & \textbf{6.6068 $\pm$ 0.0237}\\
    \hline
    \end{tabular}
    }
     \caption{Comparison of Plan2Place performance under varying hidden dimensions}
    \label{tab:vary_hidden_dim}
\end{table}
\subsubsection{Effect of the Global State Context}
\begin{figure}[ht!]
\begin{center}
   \includegraphics[width=1\linewidth]{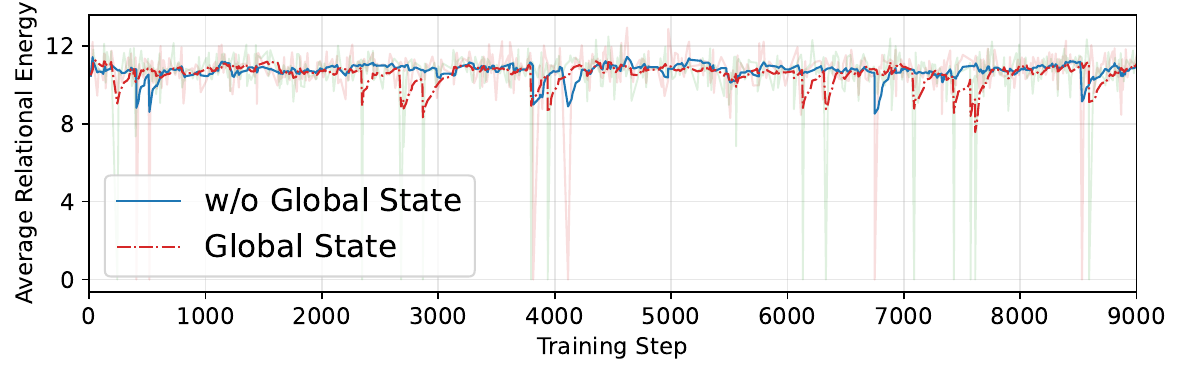}
\end{center}
   \caption{Illustration of the impact of global context state on the average relational energy during learning. The red curve (with global state context) exhibits slightly lower average $E_{rel}$ compared to the blue curve (without global state context), demonstrating the effectiveness of incorporating global contextual information in mitigating constraint violations.}
\label{fig:compare_state}
\end{figure}

Fig \ref{fig:compare_state} illustrates the impact of the global state context on constraint-aware reward learning. By utilizing the global state as the query and the local state as the keys and values within the cross-attention mechanism, the model conditions local object features on the overall scene structure. This design effectively captures the interplay between global layout constraints and local object relationships, resulting in improved satisfaction of constraint-based rewards.

\subsection{Prompts}
\label{bbox:prompt_room_classer}
\begin{tcolorbox}[height=5.5cm, valign=top]
\scriptsize
    You are a room classifier. \\
    Given a list of possible room types and the objects inside a room,
    predict the most likely room type. \\
    
    Examples: \\
    Possible rooms: [bedroom, bathroom, living room, kitchen] \\
    Objects: [bed, pillow, nightstand] $\rightarrow$ Answer: bedroom \\
    Objects: [toilet, bathtub, towel] $\rightarrow$ Answer: bathroom \\
    Objects: [sofa, TV, coffee table] $\rightarrow$ Answer: living room \\
    Objects: [stove, refrigerator, sink] $\rightarrow$ Answer: kitchen \\

    Provide a concise response, omitting any additional text at the beginning or end. \\
    Now classify this: \\
    Possible rooms: \{room\_types\} \\
    Objects: \{objects\} \\
    Answer: \\
\end{tcolorbox}

\subsubsection{More Generated Results}
\label{label:visualization_quality}
\begin{figure}[t]
\centering
\includegraphics[scale=0.23]{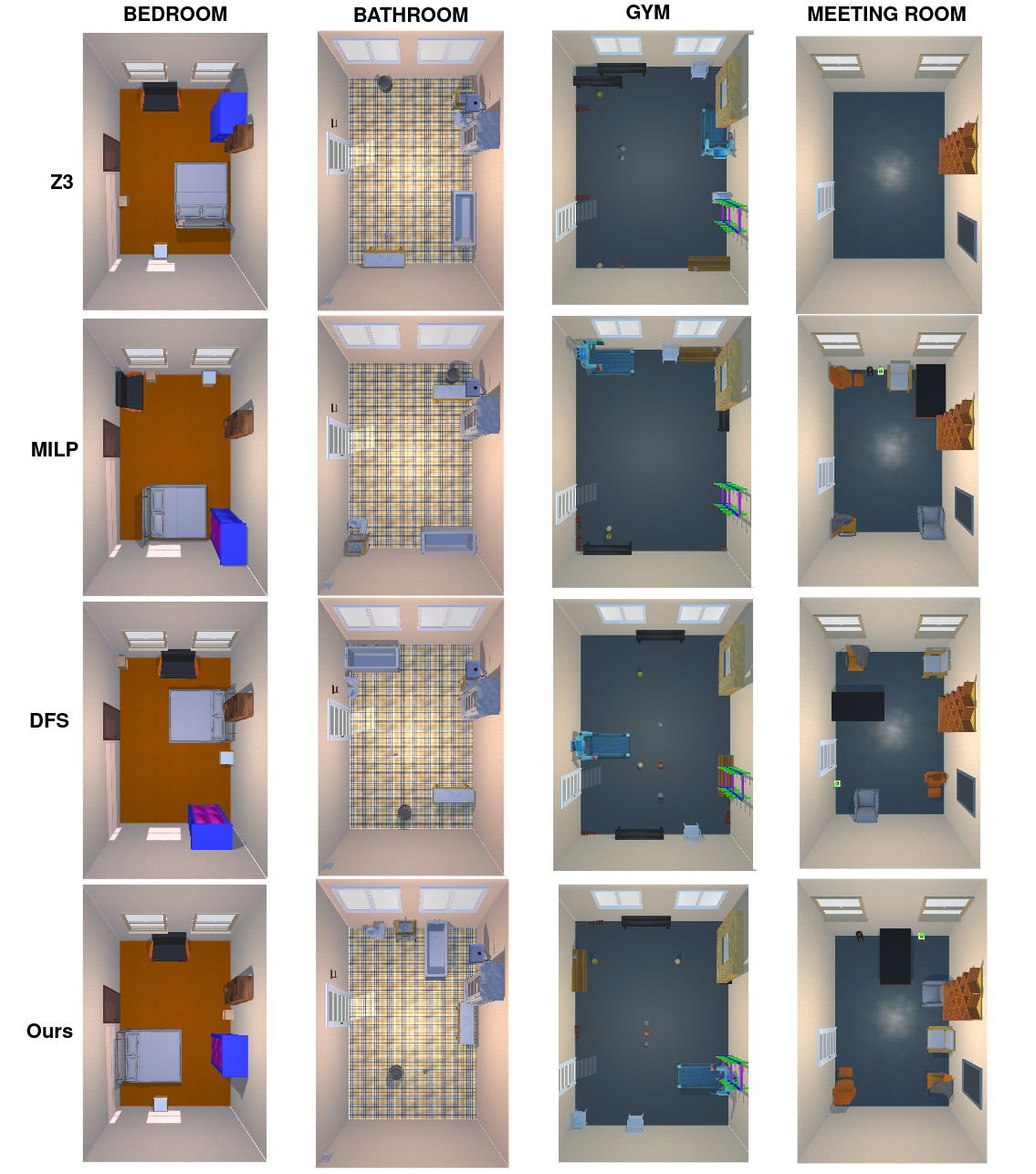}
\caption{Qualitative comparison of the different object placement methods.}
   \label{fig:compare_scene}
\end{figure}

\leavevmode\\
\textbf{- Additional Qualitative Comparison with Baseline Methods}
Fig \ref{fig:compare_scene} illustrates a qualitative comparison of different object placement methods, including Z3, MILP, DFS, and Plan2Place (ours).

The first two columns present 3D scenes generated for a bedroom and a bathroom, both of which are included in our dataset.
For the bedroom, Z3 and MILP produce layouts with reasonable coherence; however, the bed is placed too far from the wall. This issue is addressed by both DFS and Plan2Place. DFS tends to cluster objects on the left side of the door, whereas our method distributes objects more evenly throughout the room. In our layout, objects are arranged around the space, with the bed positioned and chair close to the walls and away from the main door. This leads to improved layout coherence, and object accessibility.
For the bathroom, the toilet in the layouts generated by Z3 and MILP is placed facing walls or other objects (e.g., the sink), making it difficult to use. Although DFS resolves this issue, it places the toilet near the entrance, which may reduce usability. In contrast, Plan2Place positions the toilet near the sink and away from the entrance, resulting in a more realistic layout and improved accessibility.

The last two columns illustrate a gym and a meeting room, which are not included in our dataset. The layouts generated by Plan2Place maintain coherence, realism, and object accessibility (e.g., chairs are arranged toward the center of the room), showing more consistent and reasonable arrangements than DFS and MILP in these unseen settings. In contrast, Z3 fails to find a feasible object placement solution.

The results presented in Section \ref{label:main_plan2place} and Fig \ref{fig:compare_scene} indicate that Plan2Place generates layouts that outperform the baseline methods in most cases, producing more coherent, practical, and well-structured arrangements. These findings suggest that Plan2Place has strong potential for application in 3D scene generation.

\begin{figure}[t]
\begin{subfigure}{\linewidth}
\centering
    \includegraphics[scale=0.26]{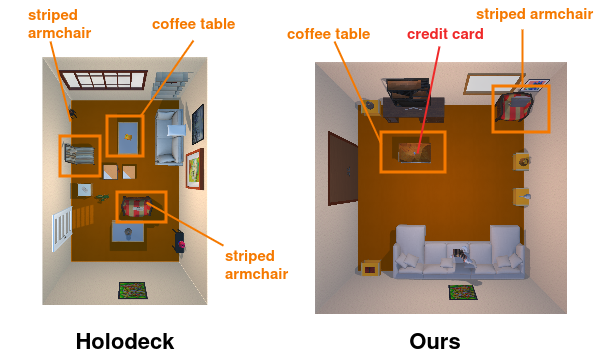}
    \caption{\textit{a living room}, and \textit{Move a {\color{red}{credit card}} from the {\color{orange}{coffee table}} to the {\color{orange}{striped armchair}}}}
    \label{fig:sample1}
\end{subfigure}
\begin{subfigure}{\linewidth}
\centering
    \includegraphics[scale=0.26]{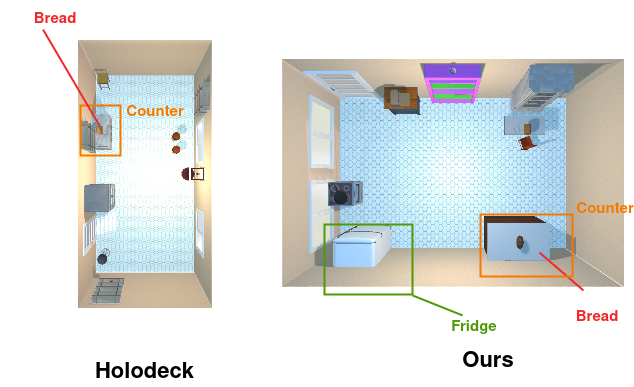}
    \caption{\textit{a kitchen}, and \textit{Put {\color{teal}{chilled}} {\color{red}{bread}} on the {\color{orange}{counter}}}}
    \label{fig:sample2}
\end{subfigure}
 \caption{Visualization of 3D scenes generated by Holodeck and our method.}
   \label{fig:compare_holodeck_scene}
\end{figure}
\textbf{- Further Qualitative Comparisons with Holodeck}
As illustrated in Fig \ref{fig:sample1}, we generate a 3D living room scene for the task: move a credit card from the coffee table to the striped armchair. Holodeck generates key objects such as the striped armchair and coffee table but fails to include the credit card. This omission results in an incomplete task due to the absence of necessary objects. In contrast, our method ensures that all key objects are present in the generated scene, making it more suitable for task-aware embodied environments.
In Fig \ref{fig:sample2}, the input specifies a kitchen scenario with the task: place chilled bread on the counter. Holodeck fails to generate the implicit object “fridge,” which is required to perform the “chill” action before placing the bread on the counter. Our method addresses this limitation by generating all necessary objects, including the bread, fridge, and counter, enabling successful task completion.


\end{document}